\documentclass[10pt,twocolumn,letterpaper]{article}

\PassOptionsToPackage{table,dvipsnames}{xcolor}
\usepackage{style/cvpr}
\usepackage{times}
\usepackage{graphicx}
\usepackage{xspace}
\usepackage{comment}
\usepackage{soul}
\usepackage{pifont}
\usepackage{xparse}

\usepackage{epsfig}
\usepackage{wrapfig}
\usepackage{subcaption}
\usepackage{dblfloatfix}
\usepackage[font=small,labelfont=bf]{caption}

\usepackage{amsmath, amsthm, amssymb}
\usepackage{textcomp}
\usepackage{stmaryrd}
\usepackage{upgreek}
\usepackage{bm}
\usepackage{cases}
\usepackage{mathtools}
\usepackage{arydshln}
\usepackage{multirow}

\usepackage{paralist}

\usepackage{cite}
\usepackage{bbm}

\usepackage{multirow}
\usepackage{rotating}
\usepackage{booktabs}
\usepackage[table]{xcolor}
\usepackage{tabularx}

\usepackage{url}
\usepackage{xspace}
\usepackage[T1]{fontenc}

\newcommand{\supp}[1]{\sethlcolor{green}\hl{#1}}

\renewcommand{\supp}[1]{#1}

\newcommand{\secref}[1]{Sec.~\ref{#1}\xspace}

\newcommand{\vilbert}{ViLBERT\xspace}
\newcommand{\mtvilbert}{MT-ViLBERT\xspace}
\newcommand{\stvilbert}{ST-ViLBERT\xspace}
\newcommand{\intravilbert}{$\mbox{MT-ViLBERT}_{\mbox{\small INTRA}}$\xspace}
\newcommand{\wogfourvilbert}{$\mbox{MT-ViLBERT}_{\mbox{\small w/o G4}}$\xspace}
\newcommand{\mtvilbertft}{$\mbox{MT-ViLBERT}_{\mbox{\small FT}}$\xspace}

\newcommand{\xhdr}[1]{\vspace{0pt}\noindent\textbf{#1}}

\renewcommand{\mtvilbert}{All-Tasks (AT)\xspace}
\renewcommand{\stvilbert}{Single-Task (ST)\xspace}
\renewcommand{\intravilbert}{Group-Tasks (GT)\xspace}
\renewcommand{\wogfourvilbert}{All-Tasks$_{\mbox{\small w/o G4}}$\xspace}
\renewcommand{\mtvilbertft}{Single-
Task$_{\mbox{~FT}}$\xspace}
\newcommand{\dynamicStartStop}{dynamic stop-and-go}
\newcommand{\DynamicStartStop}{Dynamic stop-and-go}

\renewcommand{\mtvilbertft}{AT $\xrightarrow{\small\mbox{finetune}}$ST\xspace}
\newcommand{\intravilbertft}{GT $\xrightarrow{\small\mbox{finetune}}$ST\xspace}

\newcommand{\jiasen}[1]{{\color{black}{#1}}}

\belowdisplayskip \abovedisplayskip
\newcommand{\csection}[1]{
    \vspace{-0.11in}
    \section{#1}
    \vspace{-0.10in}
}

\newcommand{\csubsection}[1]{
    \vspace{-0.09in}
    \subsection{#1}
    \vspace{-0.08in}
}

\usepackage[pagebackref=true,breaklinks=true,colorlinks,bookmarks=false]{hyperref}
\usepackage[ruled,vlined]{algorithm2e}
\usepackage{algpseudocode}

\cvprfinalcopy % *** Uncomment this line for the final submission

\def\cvprPaperID{2172} % *** Enter the CVPR Paper ID here

\ifcvprfinal
\fi
\begin{document}

%%%%%%%%% TITLE
\title{12-in-1: Multi-Task Vision and Language Representation Learning}

\author{Jiasen Lu$^3$\thanks{Equal contribution} \quad Vedanuj Goswami$^1$\footnotemark[1] \quad Marcus Rohrbach$^{1}$ \quad Devi Parikh$^{1,3}$ \quad Stefan Lee$^{2}$ \\
$^1$Facebook AI Research \quad $^2$Oregon State University \quad$^3$Georgia Institute of Technology\\
\tt\small \{vedanuj, mrf\}@fb.com \tt\small leestef@oregonstate.edu \tt\small \{jiasenlu, parikh\}@gatech.edu}

% \and
\maketitle
\thispagestyle{empty}
\begin{abstract}
\vspace{-0.15in}

Much of vision-and-language research focuses on a small but diverse set of independent tasks and supporting datasets often studied in isolation; however, the visually-grounded language understanding skills required for success at these tasks overlap significantly. 
In this work, we investigate these relationships between vision-and-language tasks by developing a large-scale, multi-task training regime. Our approach culminates in a single model on 12 datasets from four broad categories of task including visual question answering, caption-based image retrieval, grounding referring expressions, and multi-modal verification. Compared to independently trained single-task models, this represents a reduction from approximately 3 billion parameters to 270 million while simultaneously improving performance by 2.05 points on average across tasks. We use our multi-task framework to perform in-depth analysis of the effect of joint training diverse tasks. Further, we show that finetuning task-specific models from our single multi-task model can lead to further improvements, achieving performance at or above the state-of-the-art.

\end{abstract}
\vspace{-0.15in}
\csection{Introduction}

%Introduce problem
A compelling reason to study language and vision jointly is the promise of language as a universal and natural interface for visual reasoning problems -- useful both in specifying a wide range of problems and in communicating AI responses. However, the current research landscape for visually-grounded language understanding is a patchwork of many specialized tasks like question answering or caption generation, each supported by a handful of datasets. As such, progress in this field has been measured by the independent improvement of bespoke models designed and trained for each of these specific tasks and datasets. 

The recent rise of general architectures for vision-and-language  \cite{lu2019vilbert, tan2019lxmert, li2019visualbert, alberti2019fusion, li2019unicoder, su2019vl, zhou2019unified} reduces the architectural differences across tasks. {These models pretrain common architectures on self-supervised tasks to learn general visio-linguistic representations then fine-tune for specific datasets; however, the result is still a menagerie of independent task-specific models rather than a single unified model.}  This is dissatisfying in practice -- the model that understands questions cannot ground noun phrases, the grounding model cannot retrieve images based on a description, and so forth. Further, this approach does not scale well as each new task requires storing a new model. 

\begin{figure}[t]
\centering
\includegraphics[width=\columnwidth]{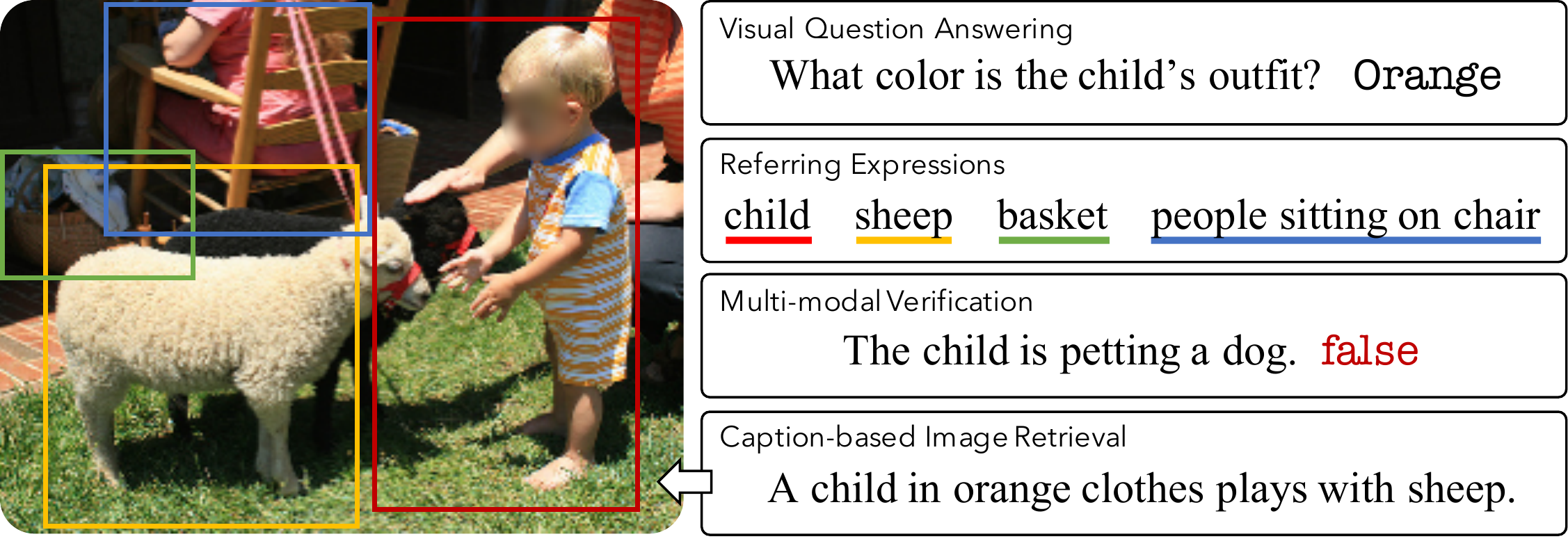}\vspace{3pt}
\caption{We introduce an approach for effective multi-task learning, training a single model on 12 popular vision-and-language datasets. This single model performs at par or even better than independent task-specific state-of-the-art approaches for many tasks.
% We develop a multi-task model trained on 12 popular vision-and-language datasets -- enabling a single model to perform many tasks at par or even better than independent task-specific state-of-the-art approaches.
}
\vspace{-8 pt}
\label{fig:model}
\end{figure}

%Talk about possible benefits
Beyond being intellectually dissatisfying, this task-based fracturing %of visually-grounded language understanding 
leaves quite a lot on the table. While individual tasks present different challenges and diverse interfaces, the underlying associations between language and visual concepts are often common across tasks. For example, learning to ground the referring expression ``small red vase'' requires understanding the same concepts as answering the question ``What color is the small vase?''. Training multiple tasks jointly can potentially pool these different sources of grounding supervision. Further, developing models that can perform well on a wide range of tasks simultaneously can help guard against the research community overfitting to specific datasets and metrics. 

%What do we do?
In this work, we develop a multi-task model for discriminative vision-and-language tasks based on the recently proposed ViLBERT\cite{lu2019vilbert} model. We consider four categories of tasks -- training jointly on a total of 12 different datasets. Our results not only show that a single model can perform all these tasks, but also that joint training can 
%lead to improvements on
\jiasen{improve the performance}
%task metrics 
compared to single-task training with the same architecture. {Before undertaking this effort, it was not obvious to us that this would be the case -- multitask training is \jiasen{notoriously} challenging and vision-and-language datasets vary greatly in size, interface, and difficulty.}
Our model attains improvements of 0.25 to 4.19 absolute points from multi-task training -- improving over corresponding single-task models for 11 out of 12 tasks. Further, we demonstrate that multi-task training is an effective pretraining step for single-task models -- leading to further gains and setting a new state-of-the-art for 7 out of 12 tasks.

{Large-scale multi-task learning is challenging as datasets can vary in size and difficulty.
To address these issues, we introduce a dynamic stop-and-go training scheduler, task-dependent input
tokens, and simple hyper-parameter heuristics. Using our proposed pipeline, we were
able to train many multi-task models with varying datasets -- assessing the relationships 
between different vision-and-language tasks in terms of their performance when trained together.}

%Summary time!
\noindent To summarize, we make the following contributions:
\begin{compactitem}[\hspace{3pt}--]
\item  %\marcus{maybe add more claim the clean dataset, could also go as first point in contributions:} 
We systematically analyze the joint training relationships between different of vision-and-language datasets and tasks and present a \emph{Clean V\&L Multi-Task setup}, which ensures no train-test leaks across task. %We systematically analyze the joint training relationships between different of vision-and-language datasets and tasks -- providing insight for future work.
\item We develop a single multi-task model trained on \textbf{12} popular V\&L datasets. Compared to a set of independent models, this represents a reduction from $\sim$3 billion parameters to $\sim$270 million while simultaneously \emph{improving} average performance by 2.05 points. 
%\marcus{I think we should always separate two aspects: (1) multi-task single-model and parameter discussion (2) fine-tuning to task + sota; I did that in the abstract}\\[-8pt]
\item We demonstrate that multi-task training is useful even in cases where single-task performance is paramount. On average, fine-tuning from our multi-task model for single tasks resulted in an average improvement of 2.98 points over baseline single-task trained models. \\[-8pt]
\end{compactitem}

\csection{Vision-and-Language Tasks}
\vspace{5pt}

\csubsection{Task-Groups and Datasets}
\label{sec:task_group}
We consider 12 popular vision and language datasets.
 These datasets cover a wide range of tasks %literature 
and require diverse grounding granularity and reasoning skills. We %divide
group related datasets into four groups to facilitate our analysis:

\xhdr{Vocab-based VQA.} Given an image and a natural-language question, select an answer from a fixed vocabulary. We consider three popular datasets for this group -- VQAv2\cite{goyal2017making}, GQA \cite{hudson2019gqa}, and Visual Genome (VG) QA \cite{krishna2017visual}.

\xhdr{Image Retrieval.} Given a caption and a pool of images, retrieve the target image that is best-described by the caption. We consider COCO\cite{cococaption} and Flickr30K\cite{plummer2015flickr30k} captioning datasets for this task-group.

\xhdr{Referring Expressions.} Given a natural language expression and an image, identify the target region that is referred to by expression. The expression can vary greatly across datasets from simple noun phrases to multi-round dialogs. We consider phrase grounding in RefCOCO(+/g) \cite{kazemzadeh2014referitgame,mao2016generation}, Pointing questions in Visual7W \cite{zhu2016visual7w}, and dialog sequences in the GuessWhat \cite{de2017guesswhat}. We note that these language inputs vary significantly in terms of detail and structure.

\xhdr{Multi-modal Verification.} Given one or more images and a natural language statement, judge the correctness or predict their semantic relationship. We consider NLVR$^2$ \cite{suhr2019corpus} and SNLI-VE \cite{xie2018visual}. In NLVR$^2$, two images are given and the statement must be true for both to be true. In SNLI-VE, image-statement pairs are classified as representing an entailment, contradiction, or neutral. That is, whether the content of the image confirms, refutes, or is insufficient to comment on the truth of the corresponding statement. 

\begin{table}[t]
\centering
\setlength{\tabcolsep}{3.5pt}
\renewcommand{\arraystretch}{0.95}
\resizebox{\columnwidth}{!}{
\begin{tabular}{l c c c c c c c c c c c c}
\toprule
	& \multicolumn{12}{c}{\%  Row-Task Test Images in Column-Task Train/Val Set}\\
	& \texttt{[A]}& \texttt{[B]}& \texttt{[C]}& \texttt{[D]}& \texttt{[E]}& \texttt{[F]}& \texttt{[G]}& \texttt{[H]}& \texttt{[I]}& \texttt{[J]}& \texttt{[K]}& \texttt{[L]}\\\midrule
	\texttt{[A]} VQA2.0\cite{goyal2017making}	&             0\%& 	0\%	& 0\%& 	0\%	& 0\%& 	0\%	& 0\%& 	0\%& 	0\%& 	0\%& 	0\%& 	0\%\\
	\texttt{[B]} VG QA\cite{krishna2017visual}	&         0\%& 	0\%	& 0\%& 	0\%	& 0\%& 	0\%	& 0\%& 	0\%	& 0\%& 	0\%	& 0\%& 	0\%\\
	\texttt{[C]} GQA\cite{hudson2019gqa}	     &            0\%& 	0\%& 	0\%& 	0\%& 	0\%& 	0\%& 	0\%& 	0\%& 	0\%& 	0\%& 	0\%& 	0\%\\
	\texttt{[D]} COCO\cite{cococaption}	&             \cellcolor{gray!100} 100\%	& \cellcolor{gray!43} 43\%& 	\cellcolor{gray!33}33\%& 	0\%& 	0\%	& 0\%& 	0\%	& 0\%& 	\cellcolor{gray!7}7\%& 	\cellcolor{gray!46}46\%	& 0\%	& 0\%\\
	\texttt{[E]} Flickr30k\cite{plummer2015flickr30k}	 &        0\%& 	0\%	&0\%&	0\%&	0\%&	0\%&	0\%&	0\%&	0\%&	0\%&	\cellcolor{gray!98}98\%& 0\%\\
	\texttt{[F]} RefCOCO\cite{kazemzadeh2014referitgame}	   &         \cellcolor{gray!100}100\%&	\cellcolor{gray!36}36\%&	\cellcolor{gray!27}27\%&	\cellcolor{gray!100}100\%&	0\%&	0\%&	0\%&	\cellcolor{gray!66}66\%&	\cellcolor{gray!8}8\%&	\cellcolor{gray!62}62\%&	0\%&	0\% \\
	\texttt{[G]} RefCOCO+\cite{kazemzadeh2014referitgame}	  &      \cellcolor{gray!100}100\%&	\cellcolor{gray!38}38\%&	\cellcolor{gray!27}27\%&	\cellcolor{gray!100}100\%&	0\%&	0\%&	0\%&	\cellcolor{gray!66}66\%&	\cellcolor{gray!8}8\%&	\cellcolor{gray!62}62\%&	0\%&	0\% \\
	\texttt{[H]} RefCOCOG \cite{mao2016generation}	&        \cellcolor{gray!100}100\%&	\cellcolor{gray!41}41\%&	\cellcolor{gray!31}31\%&	\cellcolor{gray!100}100\%&	0\%&	\cellcolor{gray!53}53\%&	\cellcolor{gray!53}53\%&	0\%&	\cellcolor{gray!8}8\%&	\cellcolor{gray!63}63\%&	0\%&	0\% \\
	\texttt{[I]} Visual 7W \cite{zhu2016visual7w}	&        \cellcolor{gray!50}50\%&	\cellcolor{gray!100}100\%&	\cellcolor{gray!79}79\%&	\cellcolor{gray!48}48\%&	0\%&	\cellcolor{gray!8}8\%&	\cellcolor{gray!8}8\%&	\cellcolor{gray!10}10\%&	0\%&	\cellcolor{gray!24}24\%&	0\%&	0\% \\
	\texttt{[J]} GuessWhat\cite{de2017guesswhat}	 &       \cellcolor{gray!100}100\%&	\cellcolor{gray!40}40\%&	\cellcolor{gray!31}31\%&	\cellcolor{gray!96}96\%&	0\%&	\cellcolor{gray!20}20\%&	\cellcolor{gray!20}20\%&	\cellcolor{gray!26}26\%&	\cellcolor{gray!7}7\%&	0\%&	0\%&	0\% \\
	\texttt{[K]} SNLI-VE\cite{xie2018visual}	& 0\%&	    0\%&	0\%&	0\%&	\cellcolor{gray!94}94\%&	0\%&	0\%&	0\%&	0\%&	0\%& 0\%&	0\% \\
	\texttt{[L]} NLVR$^2$ \cite{suhr2019corpus}	  &          0\% &	    0\%&	0\%&	0\%&	0\%&	0\%&	0\%&	0\%&	0\%&	0\%&	0\%&	0\%\\\bottomrule
\end{tabular}}
\smallskip
\caption{Percentage of row-task test images that are present in column-tasks train/val images. 
\vspace{-8 pt}
}
\label{tab:overlap}
\end{table}

\csubsection{A Clean V\&L Multi-Task Setup} %Quantifying Dataset Overlap} 
\label{sec:overlap}
Many V\&L tasks are built on top of each other and share significant overlap in terms of individual images. 

However, as each task is often examined in isolation, there does not exist an in-depth analysis of this overlap across different V\&L tasks. Table~\ref{tab:overlap} shows the percentage of test images for the target tasks which are present in other tasks' train/val sets. As we can see, there exists significant overlap across tasks. 
Even though different tasks require different inputs and outputs, other task annotations will provide clues about the visual grounding -- for example, a referring expression for a ``blue striped ball'' at training could unfairly improve a VQA model's ability to answer ``What color is the striped ball?'' for the same image at test time. 
To avoid information leakage from the annotations of other tasks, we propose a \textit{cleaned} multi-task split for V\&L tasks where test images are removed from train/val for all the tasks. {We stress that the test sets are not modified in any way, so our results are comparable to prior work.}
Cleaning results in about an 11\% reduction in training data on average across datasets. \supp{Full details of this process and statistics regarding cleaned dataset size are available in the supplement.} 

\csection{Approach}
\label{sec:approach}
\vspace{5pt}

\csubsection{Base Architecture}
\label{sec:preliminaries}
There has been a flurry of recent work developing general vision-and-language model architectures that are amenable to large-scale self-supervised pretraining. \cite{lu2019vilbert, tan2019lxmert, li2019visualbert, alberti2019fusion, li2019unicoder, su2019vl, zhou2019unified}. By pretraining general representations and then finetuning on single downstream tasks, these models set state-of-the-art in many tasks. For the base architecture in our experiments, we take the ViLBERT model proposed by Lu \etal \cite{lu2019vilbert}. We describe it here briefly.

At the interface level, ViLBERT takes as input an image $I$ and text segment $Q$ represented as the sequence $\{$\texttt{IMG}$, v_1, \dots, v_\mathcal{T},$ \texttt{CLS}, $w_1, \dots, w_T,$ \texttt{SEP}$\}$ where $\{v_i\}_{i=1}^\mathcal{T}$ are image region features \cite{anderson2018bottom}, $\{w_j\}_{j=1}^\mathcal{T}$ are word tokens, and the \texttt{IMG}, \texttt{CLS}, and \texttt{SEP} tokens are special markers. The model then outputs embeddings for each input $\{h_{v_i}\}_{i=1}^\mathcal{T}$, $\{h_{w_j}\}_{j=1}^T$,  $h_{\mbox{\texttt{IMG}}}$, $h_{\mbox{\texttt{CLS}}}$, and $h_{\mbox{\texttt{SEP}}}$. As in \cite{lu2019vilbert}, we take $h_{\mbox{\texttt{IMG}}}$ and $h_{\mbox{\texttt{CLS}}}$ as holistic image and text representations.

Internally, ViLBERT consists of two parallel BERT-style \cite{devlin2018bert}
models operating over image regions and text segments. Each stream is a series of transformer blocks (TRM) \cite{vaswani2017attention} connected by co-attentional transformer layers (Co-TRM) which enable information exchange between modalities.  We use the default parameter setting, which has 6 / 12 layers of TRM for visual / linguistic streams respectively.  

Like many of the models of this class, ViLBERT is pretrained on the Conceptual Caption dataset \cite{sharma2018conceptual} with two `proxy' tasks: \textit{masked multi-modal modelling} and \textit{multi-modal alignment prediction}. The first randomly 
masks approximately 15\% of both words and image tokens and reconstructs them given the remaining inputs. The later tasks the model with predicting whether an image and caption correspond or not. After pretraining, the model can be finetuned for strong performance for various downstream tasks.

We make two important modifications to this pretraining process. First, when masking visual regions we also mask other regions with significant overlap (> 0.4 IoU) to avoid leaking visual information. This forces the model to rely more heavily on language to predict image content. Second, we do not enforce the masked multi-modal modelling loss when sampling a negative (unmatching) caption for multi-modal alignment prediction. This will effectively remove the noise introduced by negative samples. While orthogonal to our primary contribution of multi-task learning, we found these modifications to make the baseline model more effective.  \supp{For further discussion, see the supplemental material.}
All models we present are first pretrained in this manner.

\csubsection{Multi-Task Learning}

\iffalse
\begin{figure}[t]
\centering
\includegraphics[width=\columnwidth]{figures/Fig2.pdf}
\caption{Architecture of the our model for V\&L multi-task learning. We augment the input query with a task token to learn the task-aware feature embedding. }
\vspace{-5 pt}
\label{fig:model}
\end{figure}
\fi

% \csubsection{Training on Multiple Tasks}
\label{sec:mt-vilbert}

We consider a simple multi-task model where each task has a task-specific `head' network that branches off a common, shared `trunk' ViLBERT model. % (see Figure \ref{fig:model}). 
As such, we learn shared trunk parameters $\theta_s$ and a set of task-specific layers $\{\theta_t\}_{t=1}^\mathcal{T}$ for $\mathcal{T}$ tasks. Our goal is to learn parameters $\theta_s \cup \{\theta_t\}_{t=1}^\mathcal{T}$ that minimize loss across all tasks. Details on heads and other modifications follow.

\xhdr{Task Token.} While relying on the same groundings, different tasks may still require the model to process inputs differently -- \eg  referring expressions just require grounding while VQA must follow grounding with additional reasoning. To enable this, we augment the query with a task token $\texttt{TASK}_t$ such that the new input format is $\{\texttt{IMG}, v_1, \dots, v_n, \texttt{CLS}, \texttt{TASK}_t, w_1, \dots, w_m, \texttt{SEP}\}$. The architecture can then leverage this task information in a bottom-up manner. In what follows, we describe the task-specific heads by task groups.

%- Task specific decoder. 
\xhdr{Vocab-Based VQA Output:}
We compute an overall image-query representation as an element-wise product between the holistic $h_{\texttt{IMG}}$ and $h_{\texttt{CLS}}$ representations. %, which are the holistic representations of the visual and linguistic inputs respectively.  
As in \cite{anderson2018bottom, hudson2019gqa}, we treat vocab-based VQA as a multi-label
classification task -- assigning a soft target score to each answer based on its relevancy to the ground truth answer. 
We compute scores for a set of the pre-defined answers ${A}$ by using a two-layer MLP on top of the overall representation:
\begin{equation}
    P_{v}({A} \vert {I}, {Q}) = \sigma (\texttt{MLP}({h}_{\texttt{IMG}} \odot {h}_{\texttt{CLS}}))
\end{equation}
where $\sigma$ is the sigmoid function. Due to the answer vocabulary differences, VQA and VG QA share the MLP and answer vocabulary while GQA learns a separate one.   

\xhdr{Image Retrieval Output:} 
Using the same overall representation, we compute an alignment score between image-caption pairs as:
\begin{equation}
   \texttt{Rel}({I}, {Q}) = {W}_i({h}_{\texttt{IMG}} \odot {h}_{\texttt{CLS}})
\end{equation}
where ${W}_i \in \mathbbm{R}^{d \times 1}$ is shared across COCO and Flickr30k image retrieval tasks.
% \jiasen{Do we need to mention about the training details and we do not use the online negative minining here?} \sml{breifly}
As in \cite{lu2019vilbert}, we train a 4-way multiple-choice against hard-negatives selected off-line and then fixed. Recent work has used online hard-negative mining \cite{chen2019uniter,li2019unicoder} but this is costly to compute.

\xhdr{Referring Expressions Output:} We rerank a set of region proposals \cite{yu2018mattnet} given the referring expression. We pass the final representation ${h}_{v_i}$ for each image region $i$ into a learned projection ${W}_r \in \mathbbm{R}^{d \times 1}$ to predict a matching score. 
\begin{equation}
   \texttt{Rel}({v_{i}}, {Q}) = {W}_r{h}_{vi}
\end{equation}
Note that ${Q}$ may be either a phrase, question or dialog based on different tasks (RefCOCO+/g, Visual7W, GuessWhat). ${W}_r$ is shared across all the referring expression tasks.

\xhdr{Multi-modal Verification Output:} 
Taking NLVR$^2$ as an example, the input is a concatenation of two images (${I}_0$ and ${I}_1$) and a statement ${Q}$, that the model must judge the validity of the statement given the images. We consider this a classification problem given an embedding that encodes the two image-statement pairs (${I}_0$, $Q$) and (${I}_1$, ${Q}$). The output probability is predicted by a 2-layer MLP with softmax:
% \begin{equation}
%     P_v(c \vert I_0, I_1, Q) = \texttt{softmax}(\texttt{MLP}([\bm{h}^0 ; \bm{h}^1]))
% \end{equation}
\begin{equation}
    P_v({C} \vert {I}_0, {I}_1, {Q}) = \texttt{softmax}\left(\texttt{MLP} \left(
    \begin{bmatrix}
    {h}_{\texttt{IMG}}^0 \odot {h}_{\texttt{CLS}}^0 \\
    {h}_{\texttt{IMG}}^1 \odot {h}_{\texttt{CLS}}^1     
    \end{bmatrix} \right)\right)
    % ([\bm{h}^0 ; \bm{h}^1]))
\end{equation}

\noindent where $[ \; ]$ is concatenation. 

For SNLI-VE, the input is a single image and statement. We thus learn a separate classifier of the same form that predicts the sentiment (entailment, neutral, contradiction) from the inputs.

\csubsection{Large-Scale Multitask Training}
\label{sec:training}

With 6 task heads, 12 datasets, and over 4.4 million individual training instances -- training our multi-task ViLBERT model is a daunting proposition. Multi-task learning (especially at this scale) poses significant challenges as learning objectives have complex and unknown dynamics and may compete \cite{standley2019tasks}. Further, vision-and-language datasets vary significantly in size and difficulty. For instance, a single epoch of VG (our largest dataset) corresponds to 19.8 epochs of RefCOCOg (our smallest). Likewise, when trained in isolation RefCOCOg converges in 5K iterations whereas VQAv2 takes 84K iterations (over 16 times more).  Below, we describe the details of our multi-task training approach and techniques to overcome these challenges.

\xhdr{Pretraining.} All our models are pretrained {on Conceptual Caption dataset \cite{sharma2018conceptual}} including our self-supervised task modifications as described in \secref{sec:preliminaries}.

\xhdr{Round-Robin Batch-Level Sampling.} We consider a round-robin batch-level sampling regime that cycles through each task from the beginning of multi-task training.
As such, one multi-task iteration consists of each task forwarding a batch and updating parameters in sequence.

\xhdr{Dynamic  Stop-and-Go.} As noted earlier, different tasks have different difficulties and dataset sizes. Consequentially, 
%round robin training
{simply cycling through all tasks} 
may drastically over-train smaller tasks leading to overfitting. Typically early-stopping provides a strong defense to this phenomenon; however, stopping a task in multi-task training introduces problems with \textit{catastrophic forgetting} as the base network drifts over time due to other tasks. We introduce an intuitive but effective dynamic stop and go (\texttt{DSG}) mechanism to avoid these problems. 
We monitor the validation loss $s_t$ of each task $t$, computing it once per task epoch. If performance improvement is less than 0.1\% over 2 epochs, we consider it \texttt{Converged} and shift it into \texttt{stop} mode. In \texttt{DSG} \texttt{stop} mode, a task only updates every iter-gap ($\Delta$) iterations. If validation performance degrades by 0.5\% from the task's best measured performance while in \texttt{stop} mode, the task is considered \texttt{Diverged} and is returned to \texttt{DSG} \texttt{go}. This procedure is shown in Algorithm \ref{algo}.

\begin{figure}
\centering\noindent
\hspace{-10pt}\resizebox{1\columnwidth}{!}{\centering\noindent
\csname @twocolumnfalse\endcsname
\SetKwFor{For}{for}{:}{end}
\SetKwFor{If}{if}{:}{}
\SetKwFor{uIf}{if}{:}{}
\SetKwFor{ElseIf}{else if}{:}{}
\begin{algorithm}[H]\footnotesize
\SetAlgoLined
\DontPrintSemicolon
% $\texttt{total\_iters} = \max_{\forall T \in \mathcal{T}} (  \texttt{iters}_T)$ \;
% Total Iteration $N \gets \max_{\forall T \in \mathcal{T}} (  \texttt{iters}_T) \times E $ \;
$n_t \gets$ number of iterations per epoch for task $t$ \;
$\Delta  \gets$ size of gap between iterations in  \texttt{stop} mode  \;
$\texttt{DSG}_t \gets$ $\texttt{go}$ \;
 \For {$i$ $\gets$ $1$ \KwTo MaxIter}{
 \For {$t \in \mbox{Tasks}$}{
    \If {\rm {$\texttt{DSG}_t =$} {\rm \texttt{go}} {\bf or} (\rm{$\texttt{DSG}_t = \texttt{stop}$ } {\bf and} $i\bmod \Delta = 0$)}{
    Compute task loss $L_{t}(\theta)$ and gradient $\nabla_t (\theta)$\; 
    Update $\theta \leftarrow \theta - \epsilon \nabla_t (\theta)$, where $\theta=\theta_s \cup \theta_t$\;
    }
    \If {$i \bmod n_t = 0$}{
         Compute validation score $s_t$ on task $t$\;
         \uIf {$\texttt{DSG}_t =$\texttt{go}~~{\bf and} $\texttt{Converged}~(s_t$)}{
            $\texttt{DSG}_t \gets$ $\texttt{stop}$
         }
         \ElseIf {$\texttt{DSG}_t =$\texttt{stop}~~{\bf and} $\texttt{Diverged}~(s_t$)}{
            $\texttt{DSG}_t \gets$ $\texttt{go}$
         }
    }
 }
}
\caption{\texttt{DSG} for Multi-Task Learning}
\label{algo}
\end{algorithm}}
%\end{minipage}}
\vspace{-10pt}
\end{figure}

\iffalse
\begin{algorithm}\footnotesize
\SetAlgoLined
\DontPrintSemicolon
% $\texttt{total\_iters} = \max_{\forall T \in \mathcal{T}} (  \texttt{iters}_T)$ \;
% Total Iteration $N \gets \max_{\forall T \in \mathcal{T}} (  \texttt{iters}_T) \times E $ \;
$T \gets $ \#tasks; $N \gets$ number of total iterations\;
$n_t \gets$ number of per epoch iterations for task $t$ \;
$\Delta  \gets$ size of gap between iterations in  \texttt{stop} mode  \;
$\texttt{DSG}_t \gets$ $\texttt{DSG}$ controller for task $t$, default in \texttt{go} state \;
 \For {$i$ $\gets$ $1$ \KwTo $N$}{
 \For {$t \gets$ $1$ \KwTo $T$}{
    \If {\rm {$\texttt{DSG}_t =$} {\rm \texttt{go}} {\bf or} (\rm{$\texttt{DSG}_t = \texttt{stop}$ } {\bf and} $t\bmod \Delta = 0$)}{
    Compute Loss $L_{t}(\theta)$ with respect to task $t$\;
    Compute Gradient $\nabla (\theta)$ with respect to task $t$\;
    Update $\theta \leftarrow \theta - \epsilon \nabla (\theta)$, where $\theta=\theta_s \cup \theta_t$\;
    }
    \If {$i \bmod n_t = 0$}{
         Compute validation score $s_t$ on task $t$\;
         Update controller $\texttt{DSG}_t$'s state based on $s_t$ \;
    }
 }
}
\caption{\texttt{DSG} for Multi-Task Learning}
\label{algo}
\end{algorithm}
\fi

\xhdr{Curriculum Learning.} {Inspired by prior multi-task literature \cite{bengio2009curriculum} \cite{mccann2018natural}, we experimented with both curriculum and anti-curriculum strategies based on task difficulty. Specifically, for anti-curriculum we first train on the slowest-converging task-group G1 (Vocab-Based VQA) before starting full round-robin multi-task training. Inversely for the curriculum setting we first train on our fastest-converging task-group G3 (Referring Expressions). Different from previous observation \cite{mccann2018natural, nguyen2019multi}, we found that using no curriculum leads to superior performance when combined with other strategies proposed in this section.}

\newcommand{\band}{\rowcolor{gray!10}}

\begin{table*}[ht]
\setlength\tabcolsep{3pt}
\renewcommand{\arraystretch}{1.25}
\center
  \resizebox{\textwidth}{!}{
  \begin{tabular}{c l c  c @{\hspace{2\tabcolsep}}  c c c  c @{\hspace{2\tabcolsep}}  c c  c @{\hspace{2\tabcolsep}}  c c c c c  c @{\hspace{2\tabcolsep}}  c c c c}
  \toprule
    & & & & \multicolumn{3}{c}{\small Vocab-based VQA (G1)}  & & \multicolumn{2}{c}{\small Image Retrieval (G2)} & & \multicolumn{5}{c}{\small Referring Expression (G3)}  && \multicolumn{2}{c}{\small Verification (G4)} &   & \\ 
  \cmidrule(r){5-7}
  \cmidrule(r){9-10}
  \cmidrule(r){12-16}
  \cmidrule(r){18-19}
  & &  & & VQAv2 & GQA & VG QA & & COCO & Flickr30k & & COCO & COCO+ & COCOg & V7W & GW && NLVR$^2$ & SNLI-VE & \multirow{3}{*}{\small \shortstack{\# params\\ (\# models)}} &\multirow{3}{*}{\small \shortstack{All Tasks\\ Average}} \\
    \cmidrule(r){5-7}
  \cmidrule(r){9-10}
  \cmidrule(r){12-16}
  \cmidrule(r){18-19}
  & & \small \textit{Clean}&&  test-dev & test-dev & val &  &test(R1) & test(R1) & & test & test & test & test & test && testP & test & \\
  \toprule
  %\small\texttt{0} & \vilbert\cite{lu2019vilbert} & & 70.55 & -- & -- & -- & 58.20 & -- & ?? & -- & -- & -- & -- & -- \\
 \band \small\texttt{1} & \stvilbert && & 71.82 & 58.19 & 34.38 & &65.28 & 61.14 & & 78.63 & 71.11 & 72.24 & 80.51 & 62.81 && 74.25 & 76.72 & 3B (12) & 67.25 \\
  \small\texttt{2} &\stvilbert &  \ding{51} && 71.24 & 59.09 & 34.10 & &64.80 & 61.46 & & 78.17 & 69.47 & 72.21 & 80.51 & 62.53 && 74.25 & 76.53  & 3B (12) & 67.03\\
\midrule
  \band\small\texttt{3} &\intravilbert & \ding{51}&& 72.03 & 59.60 & 36.18 & &65.06 & 66.00 & & 80.23 & 72.79 & 75.30 & 81.54 & 64.78 && 74.62 & 76.52 & 1B (4)& 68.72 \\
  \small\texttt{4} &\mtvilbert & \ding{51}&& 72.57 & 60.12 & 36.36 & & 63.70 & 63.52 & & 80.58 & 73.25 & 75.96 & 82.75 & 65.04 && 78.44 & 76.78 & \textbf{270M (1)} & 69.08 \\
  \band\small\texttt{5} &\wogfourvilbert & \ding{51}&& 72.68 & 62.09 & 36.74 & & 64.88 & 64.62 &&  80.76 & 73.60 & 75.80 & 83.03 & 65.41 && - & - & 266M (1) & - \\
\midrule
  \small\texttt{6} &\intravilbertft & \ding{51}&& 72.61 & 59.96  & 35.81 & & 66.26 & 66.98  &&  79.94 & 72.12 & 75.18 & 81.57 & 64.56 && 74.47 & 76.34 & 3B (12) & 68.81 \\
  \band\small\texttt{7} &\mtvilbertft& \ding{51}&& 72.92 & 60.48 & 36.56 & &65.46 & 65.14 & & 80.86 & 73.45 & 76.00 & 83.01 & 65.15 & & \textbf{78.87} & 76.73 & 3B (12) & 69.55 \\
  \small\texttt{8} &\mtvilbertft & & & \textbf{73.15} & \textbf{60.65} & \textbf{36.64} & &\textbf{68.00} & \textbf{67.90} &&  \textbf{81.20} & \textbf{74.22} & \textbf{76.35} & \textbf{83.35} & \textbf{65.69} && \textbf{78.87} & \textbf{76.95} & 3B (12) & \textbf{70.24} \\

  \bottomrule
  \end{tabular}}
  \smallskip
  \caption{Comparison of our multi-task models to single-task performance. We find multi-task training (rows \texttt{3-5}) provides significant gains over single-task training (rows \texttt{1-2}) while reducing the parameter count from over 3 billion to 270 million. Further,  following multi-task training by task-specific fine-tuning (rows \texttt{6-9}) further gains can be made at the cost of increased parameters.} %\ding{51} = trained on cleaned datasets only. See text for full details.}
  \vspace{-10 pt}
  \label{tab:full}
\end{table*}

\xhdr{Setting Multi-Task Hyperparameters.} We follow a simple design philosophy 
%regarding hyper-parameters 
-- identify simple heuristics based on hyper-parameters tuned for each task in single-task training. This significantly reduces the burden of searching for joint-training hyper-parameters. \jiasen{See the supplement for a full list of per task learning rates, batch sizes, and other settings. Our code has been made available\footnote{https://github.com/facebookresearch/vilbert-multi-task}}.

\noindent\emph{Batch Size:}  {For multi-task}, we keep the batch size tuned for single-task training for each task. %, \ie batch size $b_t$ for dataset $t$.

\noindent\emph{Warm-up Duration:} We found it important to set warm-up duration relative to the largest dataset. Specifically, we run linear warm-up over $\eta * N$ iterations where $N$ is the max. number of iterations taken to train any dataset in the single-task setting. %$N$ is chosen to be maximum number of iterations across all the tasks or total number of iterations for largest dataset. 
We observe significant performance degradation for harder tasks when warm-up was shorter. We set $\eta$ to 0.1 for our experiments. % \vedanuj{same as used in all our single-task training}.

\noindent\emph{Loss Scaling:} Our model has shared and task-specific parameters and we found it important to maintain separate learning rates. For the shared base model, we set the the base learning rate to
%$\mathcal{LR}_{base} = \min_{\forall t \in \mathcal{T}} ( \mathcal{LR}_{t}^{ST})$, 
the minimum over all single-task dataset parameters. 
To accommodate variable learning rates for each dataset, we scale the task loss for each dataset by the ratio of task target learning rate over base learning rate.

\iffalse
\csubsection{Finetuned from MT-ViLBERT}
\label{sec:finetune-vilbert}

For some applications, single task performance may be paramount and be worth the extra storage for a task-specific model. Even in such cases, fine-tuning from a multi-task trained model may allow the model to take advantage of additional, diverse supervision captured during multi-task training. Following \cite{liu2019multi}, we considering the case where the model is pre-trained on all tasks at once but is then fine-tuned on the individual supervised tasks. 
\fi

%Recall that multi-task learning is an approach that can incorporate more data with different supervisions. This essentially provides regularization effect via alleviating overfitting to a specific task. 

\csection{Experiments and Results}

\begin{table*}[t] 
\renewcommand{\arraystretch}{1}
\centering
\resizebox{\textwidth}{!}{
\begin{tabular}{c l @{\hspace{1.5\tabcolsep}} cccc c c c @{\hspace{1.5\tabcolsep}} cccccc c c}
    % \cmidrule(r){1-7}
    %  \cmidrule(r){9-15}
    \toprule
     &&& \multicolumn{4}{c}{Trained With}  && & \multicolumn{6}{c}{Trained With} \\
     \cmidrule(r){4-7}
     \cmidrule(r){10-15}
   &&& \multicolumn{1}{c}{G1}  & \multicolumn{1}{c}{G2} & \multicolumn{1}{c}{G3}  & \multicolumn{1}{c}{G4} & Avg. && \multicolumn{1}{c}{G1 \& G2}  & \multicolumn{1}{c}{ G1\& G3} & \multicolumn{1}{c}{ G1 \& G4}  & \multicolumn{1}{c}{ G2 \& G3} &  \multicolumn{1}{c}{ G2 \& G4} & \multicolumn{1}{c}{ G3 \& G4} & Avg. \\  
  \cmidrule(lr){2-2}
  \cmidrule(r){4-8}
  \cmidrule(r){10-16}
  \multirow{3}{*}{\rotatebox{90}{\footnotesize Relative PERF}} & G1 (VQAv2) && - & \cellcolor{green!3.8} 0.38\% & \cellcolor{green!3.8} 0.38\% & \cellcolor{red!2} -0.20\% & \cellcolor{green!1.9} 0.19\% && - & - &	- &	\cellcolor{green!6.3} 0.63\% & \cellcolor{red!0.8}	-0.08\% &	\cellcolor{green!1.8} 0.18\% &	\cellcolor{green!2.4} 0.24\% \\
  
 & G2 (Flickr30k) && \cellcolor{green!4.6} 0.46\% & - & \cellcolor{green!2.3} 0.23\% & \cellcolor{red!41.3} -4.13\% & \cellcolor{red!11.5} -1.15\% && - &\cellcolor{green!12.4}	1.24\% &	\cellcolor{green!4.9} 0.49\% &	- &	- &	\cellcolor{red!43.6} -4.36\% &	\cellcolor{red!8.8} -0.88\% \\ 
   & G3 (Visual7W) && \cellcolor{green!3.9} 0.39\% & \cellcolor{green!7.8} 0.78\% & - & \cellcolor{green!2.4} 0.24\% & \cellcolor{green!4.7} 0.47\% && \cellcolor{green!8.6} 0.86\% &	- &	\cellcolor{green!1.9} 0.19\% &	- &	\cellcolor{green!2.9} 0.29\% &	- &	\cellcolor{green!4.4} 0.44\% \\
  & G4 (NLVR$^2$) && \cellcolor{green!22.9} 2.29\% & \cellcolor{green!14.7} 1.47\% & \cellcolor{green!6.7} 0.67\% & - & \cellcolor{green!14.8} 1.48\%  && \cellcolor{green!36.9} 3.69\% & \cellcolor{green!32.2} 3.22\% & - & \cellcolor{green!27.3} 2.73\% & - & - & \cellcolor{green!32.1} 3.21\%  \\
   \cmidrule(lr){2-2}
  \cmidrule(r){4-8}
  \cmidrule(r){10-16}
   & Avg. & & \cellcolor{green!10.4} 1.04\% & \cellcolor{green!8.8} 0.88\% & \cellcolor{green!4.3} 0.43\% & \cellcolor{red!13.6}-1.36\% & - &  & \cellcolor{green!22.7} 2.27\% & \cellcolor{green!22.3} 2.23\% & \cellcolor{green!3.4} 0.34\% & \cellcolor{green!16.8} 1.68\% & \cellcolor{green!1} 0.10\% & \cellcolor{red!20.9}-2.09\% & - \\
%   \cmidrule(lr){1-1}
%   \cmidrule(r){3-7}
%   \cmidrule(r){9-15}
  \bottomrule
  \end{tabular}}
   \smallskip
\caption{Pair-wise (left) and triple-wise (right) inter-group representative task analysis. Each entry is the relative performance change from single-task training for the row-task when jointly trained with the column-task(s).}
  \vspace{-10 pt}
\label{tab:repanalysis}
\end{table*}

%We describe out experimental setup and results below.

\vspace{5pt}

\csubsection{Single-Task Performance}
To establish baseline performance for the \vilbert architecture that forms the backbone of our multi-task experiments, we first train single-task models on top of the base ViLBERT architecture (Section \ref{sec:approach}) for each of our 12 datasets. 
%As with all our models, these are finetuned after Conceptual Captions pretraining as in \cite{lu2019vilbert}.
Rows \texttt{1} and \texttt{2}  in Table~\ref{tab:full} show the performance of these models trained on the full and cleaned datasets, respectively. As expected, reducing the training set size through cleaning results in lower performance in most cases. Our improvements over the pretraining objective (Sec \ref{sec:preliminaries}) results in better downstream tasks performance (71.82 vs.~70.55 on VQA  and 61.46 vs.~58.20 on Flickr30k Recall@1). \supp{See the supplementary for full comparison.} 
Overall, our base architecture is competitive with prior work and a good starting point for multi-task learning.

\csubsection{Intra-Group Multi-task Performance}
We begin with the most intuitive multi-task setting -- jointly training tasks within the same groups. As grouped tasks are typically highly related, this is akin to some existing data augmentation practices (\eg adding Visual Genome (VG) QA data when training VQA). Note this corresponds to four separate multi-task models -- one for each group.

Table~\ref{tab:full} row \texttt{3} shows the result of intra-group multi-task training. Comparing with single-task models trained on the same data (row \texttt{2}), we see meaningful improvements of between 0.37\% (NLVR$^2$) and 4.54\% (Flickr30k retrieval) points for 11 out of 12 tasks (only SNLI-VE did not improve). Comparing to row \texttt{1}, we see that intra-group multi-task training overcomes the data-loss from cleaning with an average score of 68.72, outperforming the single-task models trained on the full datasets which have an average score of 67.25.  Further, the total number of parameters drops by a factor of 3$\times$ -- going from 12 full models to only 4.

\csubsection{Inter-Group Multi-task Performance}

\xhdr{Representative Task Analysis.} We next consider the interplay between different task-groups. For efficiency, we consider multi-task training with representative tasks from each group %\jiasen{which cover most of the images. } 
-- specifically VQA (G1), Retrieval Flickr30k (G2), Visual7W (G3), and NLVR$^2$ (G4). These were selected to maximize diversity in underlying image sources. We examine their relationships by jointly training all pairs and triplets of tasks under our multi-task training approach.

Table~\ref{tab:repanalysis} (left) shows the results of training each representative task pair. Each entry is the percent change from single-task performance for the row-task when jointly trained with the column-task. As such, the Avg.~row (bottom) shows the mean impact each column-task has on other tasks, and likewise the Avg.~column (right) shows the mean impact other tasks have on each row-task. For instance, we find that adding VQA (G1) benefits other tasks with an average improvement of +1.04\%. Interestingly, adding NLVR$^2$ (G4) degrades other tasks on average (-1.36\%) while making significant gains itself (+1.48\%). This is primarily due to a -4.13\% interaction with G2.
Table~\ref{tab:repanalysis} (right) shows all task triplets. Gains in the paired-experiments are not simply additive. In the pair-wise analysis, G3 gained +0.39\% and +0.78\% from G1 and G2 respectively.
%however, combining G3 with G1\&2 results in only +0.86\%. 
As before, G4 has some strong negative effects on other groups (-4.36\% G2 with G3 \& G4) but these effects can be regulated by other tasks (+0.49\% G2 with G1 \& G4).

\xhdr{Full Multi-task Results.} We move to our main result -- a single model trained on all 12 datasets. The results of this All-Tasks (\texttt{AT}) model are shown in Table \ref{tab:full} row \texttt{4}. This model outperforms independent single-task models trained on the same data (row \texttt{2}) for 11 out of 12 tasks and improve the average score by 2.05 points (69.08 vs.~67.03). We reiterate for emphasis, average performance \emph{improves} by 2.05 points while \emph{reducing} the number of parameters from over 3 billion to 270 million (a 12$\times$ reduction). This is also true for comparison with single-task models trained on full datasets (row \texttt{1}) by a similar margin of 1.83 points.

Our \texttt{AT} model also outperforms the Group-Task (\texttt{GT}) models (row \texttt{3}) despite having 4x fewer parameters (avg. 69.08 vs 68.72). This implies that despite their diversity, tasks across different groups can benefit from joint training.

%As this is only a single model, it does this while having \textasciitilde 4x fewer parameters than the Group-Task models.
% and \textasciitilde12x fewer than the Single Task models. Interestingly, this model performs significantly better than both ST Clean (avg. 67.03 ) as well as ST Full (avg. 67.25) model by a huge margin. 

We observed from the representative task analysis that G4 tends to have a negatively effect other groups during joint training. To validate this observation on all tasks, we train an All-Task model without G4 (row \texttt{5}). This model achieves higher avg.~score of 67.96 for G1+G2+G3 compared to the full \texttt{AT} model's 67.39. \jiasen{NLVR$^2$ (G4) presents two images per description and often
one matches while the other does not. Despite the alignment
with one image, the instance as a whole is negative.
We speculate that this supervision may interfere with the standard 
caption-image alignment objective in Flickr30k.}

\vspace{-5pt}
\csubsection{Multi-Task Learning as Pretraining}

\begin{table}[t] \footnotesize
\setlength\tabcolsep{3 pt}
\resizebox{\columnwidth}{!}{
  \begin{tabular}{l c  c c c  c c   }
  \toprule
  \multirow{3}{*}{Task} & \multirow{3}{*}{Split} &  \multirow{3}{*}{SOTA}  & \multicolumn{2}{c}{UNITER \cite{chen2019uniter}} & Ours$_{\texttt{AT}}$ & Ours$_{\texttt{AT->ST}}$ \\  
  \cmidrule(lr){4-5}
  \cmidrule(lr){6-6}
  \cmidrule(lr){7-7}
  & & & BERT$_{\texttt{B}}$ & BERT$_{\texttt{L}}$ & BERT$_{\texttt{B}}$ & BERT$_{\texttt{B}}$\\
  \midrule
   %\multirow{2}{*}{\small{VQA}} 
   VQA & test-dev & - & 72.27 & \textbf{73.24} & 72.57 & 73.15  \\
%   & test-std & \textbf{73.40} \cite{chen2019uniter} & - & -  \\
  VG QA & val & - & - & - & 36.36 & \textbf{36.64}  \\
%   \multirow{2}{*}{\small{GQA}} 
   GQA & test-dev & 60.00 \cite{tan2019lxmert}  & - & - & 60.12 & \textbf{60.65}  \\ 
%   & test-std & \textbf{63.17} \cite{hudson2019learning} & - & - \\
  \midrule
 IR COCO & test (R1) & \textbf{68.50} \cite{li2019unicoder} & - & - & 63.70 & 68.00   \\
 IR Flickr30k & test (R1) & - & 71.50 & \textbf{73.66} & 63.52 & 67.90 \\
  \midrule
  RefCOCO & test & - & 80.21 & 80.88 & 80.58 &  \textbf{81.20}  \\
  RefCOCO+ & test & - & 72.90 & 73.73 & 73.25 & \textbf{74.22}    \\
  RefCOCOg & test & - & 74.41 & 75.77 & 75.96 & \textbf{76.35}   \\
  Visual 7W & test & 72.53 \cite{hu2017modeling} & - & -  & 82.75 & \textbf{83.35}  \\
  GuessWhat & test & 61.30 \cite{de2017guesswhat} & - & - & 65.04 & \textbf{65.69}  \\
  \midrule
  NLVR$^2$ & testP & - & 77.87 &\textbf{79.50}  & 78.44 & 78.87  \\
  SNLI-VE & test & - & 78.02 &\textbf{78.98}  & 76.78  & 76.95    \\
 \midrule
  \midrule
  \shortstack{\# params\\ (\# models)} & & & \shortstack{602M\\ (7 x 86M)} & \shortstack{2.1B\\ (7 x 303M)} & \shortstack{\textbf{270M} \\ \textbf{(1 x 270M)}} & \shortstack{3B\\ (12 x 250M)} \\
  \bottomrule
  \end{tabular}}
    \smallskip
  \caption{Comparison to recent SOTA. For image retrieval (IR) COCO and Flickr we report R1 scores on the 1K test set.}
%   \vspace{5pt}
 \label{tab:sota}
\end{table}

For some applications, single task performance may be paramount and justify  storing a task-specific model. 
Even then, fine-tuning from a multi-task trained model may allow the model to take advantage of the additional, diverse supervision captured during multi-task training. 
Following \cite{liu2019multi}, %we considering the case where the model is pretrained on all tasks at once but is then fine-tuned on the individual supervised tasks.
we finetune our trained multi-task models (\texttt{GT} and \texttt{AT}) on each downstream task and show results in Table \ref{tab:full}.
Rows \texttt{6} and \texttt{7} show that finetuning from the all-task model  (\texttt{AT}) outperforms finetuning from the group-task models (\texttt{GT}) with an average score of 69.51 vs.~68.81. For comparison with our multi-task models, these are finetuned on the cleaned datasets which are 11\% smaller on average. To compare to prior work, we also finetune on the full dataset for individual tasks (Row \texttt{8}) and observe further improvements. Recall that our multi-task model was trained on cleaned data so there is no possibility of test leak here. These model  outperform single-task models without multi-task pretraining (row \texttt{1}) by a large margin (70.24 vs.~67.25 avg.~score). 

\csubsection{Comparison with Existing Work}\label{subsec:comparison}

In Table \ref{tab:sota} we compare with existing state-of-the-art. 
We draw special comparison with the recent UNITER \cite{chen2019uniter} architecture as it is similar to our base ViLBERT model.  Like ViLBERT, UNITER is a general BERT-based vision-and-language architecture pretrained through self-supervised tasks and then finetuned for each downstream task. We show two UNITER columns corresponding to their underlying BERT model -- either Base \texttt{B} or Large \texttt{L}. Our ViLBERT model uses the smaller BERT$_{\mbox{\texttt{B}}}$.
%
%
%Compared to UNITER, our ViLBERT model is pretrained on data separate from our training datasets (Conceptual Caption \cite{sharma2018conceptual}) whereas UNITER leverages both out-of-domain (Conceptual Caption \cite{}, SBU \cite{}) and task datasets (COCO \cite{}, VG \cite{}).
%
%
Our single all-task model (Ours$_{\texttt{AT}}$) achieves competitive performance to state-of-the-art task-specific models. %Compare to UNITER, Our base model is pretrained on out-of-domain dataset (Conceptual Caption \cite{}), while UNITER is pretrained on both in-domain (COCO \cite{}, VG \cite{}) and out-of-domain (Conceptual Caption \cite{}, SBU \cite{}) data. (Ours$_{\texttt{AT}}$) also trained on cleaned dataset and is capable to solve all the tasks in a single model. 
Our single-task finetuned models  (Ours$_{\texttt{AT->ST}}$) surpass state-of-the-art on 7 out of 12 tasks.

\begin{table}[t]\footnotesize
\vspace{5pt}
\setlength\tabcolsep{2.5 pt}
\centering
\resizebox{\columnwidth}{!}{
  \begin{tabular}{l c c c c c c c c }
  \toprule
  \multicolumn{1}{c}{} & \multicolumn{1}{c}{VQA} & \multicolumn{3}{c}{COCO Retrieval} & \multicolumn{3}{c}{Flickr Retrieval} & \multicolumn{1}{c}{FG}\\

  \cmidrule(r){2-2}
  \cmidrule(r){3-5}
  \cmidrule(r){6-8}
  \cmidrule(r){9-9}
  
   &  & R1 & R5 & R10 & R1 & R5 & R10 & R1 \\
  \midrule
  OmniNet \cite{pramanik2019omninet} & 55.76 & - & - & - & - & - & - & - \\
  HDC \cite{nguyen2019multi} & 69.28 & 57.40 & 88.40 & 95.60 & 56.10 & 82.90 & 89.40 & 57.39 \\
  \midrule
  Ours & \textbf{72.70} &	\textbf{65.16} & \textbf{91.00} & \textbf{96.20} &	\textbf{65.06} &	\textbf{88.66} &\textbf{	93.52} &	\textbf{64.61} \\
  \bottomrule
  \end{tabular}}
  \smallskip
  \caption{Comparison with other multi-task models. VQA score is on test-dev  and the retrieval tasks on their respective 1K test split. For Flickr Grounding (FG) we report R1 on Flickr30K test.}
    \vspace{-5 pt}
  \label{tab:comp}
\end{table}

Table \ref{tab:comp} compares our method with other recently proposed multi-modal,
multi-task learning approaches --   OmniNet \cite{pramanik2019omninet} and Hierarchical Dense Co-Attention (HDC) \cite{nguyen2019multi}. OmniNet is trained on part-of-speech tagging, image captioning, visual question answering, and video activity recognition, while HDC is trained on image caption retrieval, visual question answering, and visual grounding.  We train a multi-task model on the same tasks and cleaned datasets used in HDC \cite{nguyen2019multi}. Flickr Grounding is a new task that we include for this comparison. Our multi-task model outperforms these approaches by a large margin. 
\csection{Analysis and Ablation Study}\label{sec:ablation_analysis}

\xhdr{Ablations on task token and training strategies. }
To verify our design choices, we perform ablations for different task token granularity and  multi-task training strategies. The results are shown in Table \ref{tab:ablation}. We report average group and overall average performance. %We report the weighted average (WA) score of the group averages to determine which model is better. 
\supp{Detailed breakdown for each task can be found in supplement}. 
 
For task tokens, our default setting is with a different task token per dataset (12 total, Row \texttt{1}). We compare this with two ablations: one task token per output head (4 total, Row \texttt{2}) and no task tokens (Row \texttt{3}). We observe that task-specific tokens lead to {better performance} compared to head-based tokens (avg.~69.08 \vs~68.52) and no task tokens (avg.~69.08 \vs~68.53). 
% reason: visual grounding. 
This shows that task-aware feature embedding is useful even within the same output space; \eg per-task tokens may help differentiate noun phrases and pointing questions in Referring Expression.

For multi-task training schedule, we compare our dynamic stop-and-go (\texttt{DSG}) (Row \texttt{3}) with Curriculum (Row \texttt{5}) and Anti-Curriculum (Row \texttt{6}) approaches discussed in \secref{sec:approach}. We consider convergence rate as a measure of task difficulty. For Curriculum,  we first train tasks in G4 and then train all tasks together (easier $\longrightarrow$ harder). For Anti-Curriculum, we train G1 tasks first and then train on all tasks together (harder $\longrightarrow$ easier). Table \ref{tab:ablation} shows our dynamic stop-and-go training schedule outperforms anti-curriculum (avg.~68.52 \vs~67.98) and curriculum (avg.~68.53 \vs~67.24). Row \texttt{7} shows results of a `vanilla', round-robin training scheme with no task tokens or training scheduling. The average score of vanilla multitask is close to anti-curriculum (67.92 \vs~67.98). Consistent with prior work \cite{mccann2018natural}, performance on harder tasks (G1)  is worse compared to anti-curriculum. Our full training regime outperforms this significantly (avg.~69.08 \vs 67.92).

\NewDocumentCommand{\rot}{O{45} O{1.5em} m}{\makebox[#2][l]{\rotatebox{#1}{#3}}}%

\begin{table}[t]\footnotesize
\setlength\tabcolsep{5 pt}
\centering
\resizebox{\columnwidth}{!}{
  \begin{tabular}{c @{\hspace{-2pt}} l c c c c c c c}
  \toprule
   && \small \shortstack{Task\\ Token} &  \small \shortstack{Dynamic\\ Stop-and-Go} & G1 & G2 & G3 & G4 & \small \shortstack{All Tasks\\ Average} \\
  \midrule
  &\texttt{AT} (our) \\
\footnotesize\texttt{1} & \ \ \ token per dataset  & \checkmark & \checkmark & \textbf{56.35} &	\textbf{63.61} &	\textbf{75.52} &	\textbf{77.61 }&	\textbf{69.08}  \\
\footnotesize\texttt{2} & \ \  \ token per head   & \checkmark & \checkmark & 55.95 &	61.48 &	75.35 &	77.37 &	68.52 \\
\footnotesize\texttt{3} &  \ \  \ w/o task token & &\checkmark 	& 55.67 & 62.55	& 75.38 & 76.73 & 68.53  \\
\footnotesize\texttt{4} &  \ \  \ w/o \texttt{DSG}	& \checkmark &  &  55.50&62.92	& 75.24 & 76.31 & 68.52  \\
\footnotesize\texttt{5} &  \ \ \ w/ curriculum  &  &   & 54.68 & 61.21 & 75.19 & 76.70 & 67.24 \\
\footnotesize\texttt{6} &  \ \ \ w/ anti-curriculum &  &  & 55.82 & 59.58	& 73.69 & 75.94 & 67.98  \\
\footnotesize\texttt{7} & \ \  \ vanilla multitask &  &  & 54.09 & 61.45 & 75.28 & 76.71 & 67.92 \\
%  \midrule
%  Curriculum training &  &   & 54.68 & 61.21 & 75.19 & 76.70 & 67.24 \\
%  Anti-Curriculum training&  &  & 55.82 & 59.58	& 73.69 & 75.94 & 67.98  \\
  \bottomrule
  \end{tabular}}
  \smallskip
  \caption{Ablations on our design choices and comparison to curriculum and anti-curriculum learning multi-task approaches. }
  \vspace{-10 pt}
  \label{tab:ablation}
\end{table}

\begin{figure}[t]
\centering
\includegraphics[width=\columnwidth,  keepaspectratio]{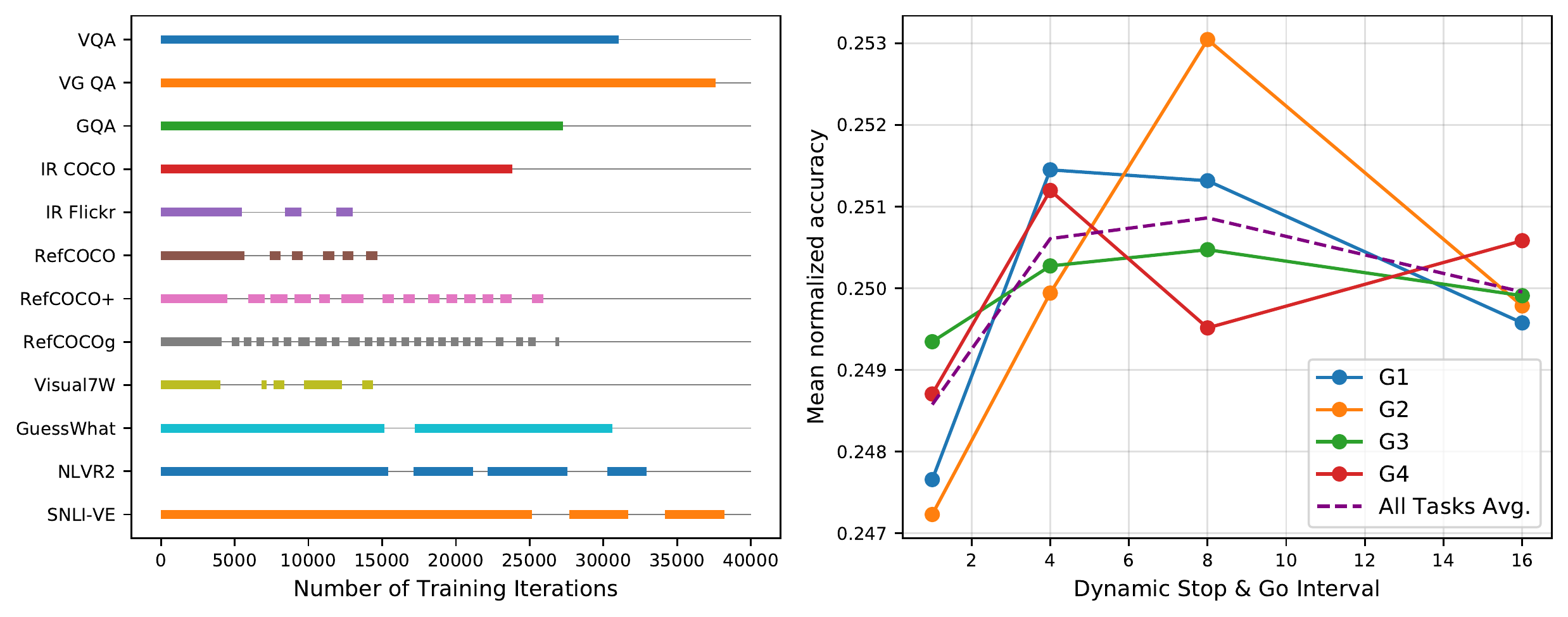}
\caption{Left: Visualization of \DynamicStartStop~during multi-task training. Solid line indicates in the \texttt{go} mode while thin line indicates \texttt{stop} mode.
Right: Mean accuracy (normalized group-wise for easier comparison)  for each group with different iter-gap $\Delta$ for \DynamicStartStop~ .}
  \vspace{-8 pt}
\label{fig:iter_gap}
\end{figure}

\xhdr{Behavior of Dynamic Stop-and-Go training. }
To characterize our dynamic stop-and-go training scheme, we visualize the dynamic training schedule in Fig.~\ref{fig:iter_gap} (left) -- bold lines indicate normal \texttt{go} training and thin lines are \texttt{stop} states when datasets receive sparser updates at a fixed iteration gap (every 4th iteration here).  We see that smaller datasets quickly converge and enter \texttt{stop} state training early. As the base model drifts over time, they periodically return to full \texttt{go} state training to adjust. Interestingly, after some cycles of this, they enter the \texttt{stop} state and continue with only sparse updates for the rest of training. %It is interesting that tasks with smaller datasets go into stop mode earlier and after some switch between stop and go, the model may learn a more general feature representations and the task remains to stay in the stop mode.  

Another aspect of \dynamicStartStop~training is the sparsity of updates in the \texttt{stop} state.  Fig.~\ref{fig:iter_gap} (right) shows the mean normalized accuracy for each group for multi-task models trained with different iteration gaps ($\Delta$). We observe that raising $\Delta$ (\ie updating more sparsely) improves performance initially but degrades for larger values. \supp{Absolute and per-task scores are provided in the supplement.}

\begin{figure*}[t]
\centering
\includegraphics[width=0.93\textwidth]{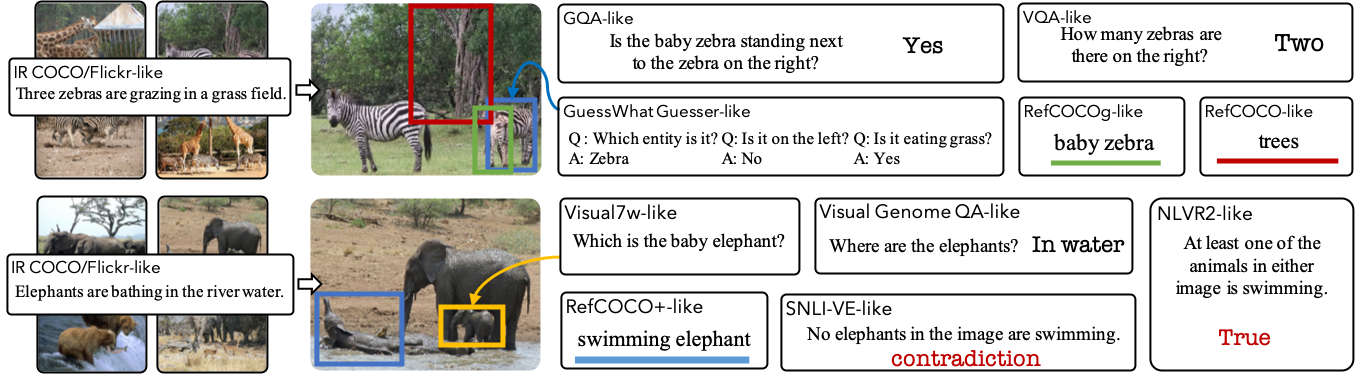}
\vspace{3pt}
\caption{
Our single model (Our$_\texttt{AT}$) can perform a multitude of V\&L tasks: caption and image retrieval, question answering, grounding phrases, guessing image regions based on a dialog, verifying facts about a pair of images, natural language inferences from an image, etc. Here we show outputs of our model for a variety of inputs (that mimic tasks from the 12 datasets it has been trained on).
}
\vspace{-10 pt}
\label{fig:qual}
\end{figure*}

\xhdr{Multi-Task visual grounding consistency. }
Given the common shared base model, one question is whether multitask models exhibit more consistent visual groundings than independent task-specific models. For example, does a model that correctly answers ``What color is the largest dog?'' also correctly ground the referring expression ``largest dog''? 
To assess this, we consider 1500 images from the RefCOCO/+ test sets that also have VQA annotations such that for each
image $I_i$ there are associated questions $\{q^{(i)}\}$ and referring expressions $\{r^{(i)}\}$.  To measure the overlap in visual concepts between a question $q^{(i)}_j$ and reference $r^{(i)}_k$, we count overlapping nouns and adjectives (identified using a part-of-speech tagger \cite{toutanova2003feature}) and denote this $d(q^{(i)}_j,r^{(i)}_k)$. Armed with this notion of similarity, we consider each question-reference pair for each image (total 111,275 combinations) and compute a weighted accuracy. A pair is considered correct if the question was answered correctly and the referent was localized. Each pair is weighed by their overlap $d(q^{(i)}_j,r^{(i)}_k)$. Note that if $q^{(i)}_j$ and $r^{(i)}_k$ do not have any common visual concept ($d(q^{(i)}_j,r^{(i)}_k)$), the correctness of this pair does not affect the overall metric.

We evaluate our Single-Task (\texttt{ST}), All-Task (\texttt{AT}), and finetuned from All-Task (\texttt{AT->ST}) models on the proposed metric. 
\texttt{AT} consistently outperforms \texttt{ST} (55.40 \% \vs~58.30\%) and \texttt{AT->ST} achieves the best performance (64.64\%). This shows our model trained on multiple tasks achieve better visual grounding consistency across different tasks. \supp{Further analysis can be found in the supplement.}

\begin{figure}[t]
\centering
\includegraphics[width=\columnwidth]{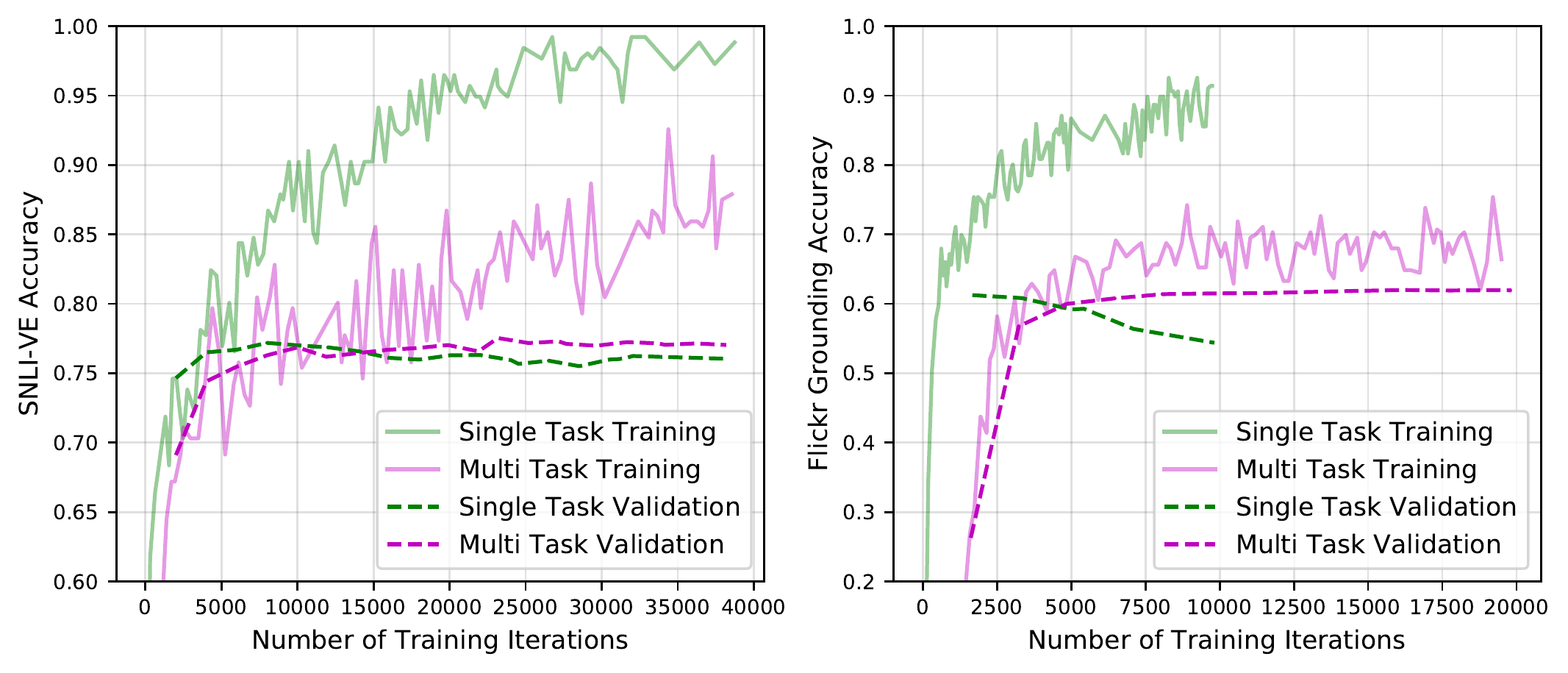}
\caption{Multi-Task training acts as a regularizer.}
  \vspace{-10 pt}
\label{fig:regularizer}
\end{figure}

\xhdr{Regularizing effects of multi-task learning.} 
We find multi-task training to have a regularizing effect on tasks which overfit when trained separately. In Fig.~\ref{fig:regularizer} we plot the training and validation curves for two tasks (SNLI-VE and Flickr Grounding) where single task training overfits quickly. On the other hand when trained in a multi-task setup with all other tasks, the validation score improves and there is no overfitting.

\xhdr{Qualitative examples.} Figure \ref{fig:qual} shows example outputs of our models. Due to space limitation, we provide extensive visualizations in the supplement.

\csection{Related Work}

% [Add related work]

\xhdr{Multi-task learning.} 
There has been substantial interest in multi-task learning \cite{caruana1997multitask, ruder2017overview}, \ie training a single model for multiple tasks at once.  Advances in multi-task learning have been developed in the context of vision \cite{zhang2013robust, zhang2014facial, misra2016cross, kokkinos2017ubernet, strezoski2019many, bragman2019stochastic}, language \cite{collobert2008unified, liu2015representation, mccann2018natural, liu2019multi, raffel2019exploring}, 
and robotics \cite{parisotto2015actor, jaderberg2016reinforcement, teh2017distral}. 
% some paper we should mention about:
Among them, 
Standley \etal \cite{standley2019tasks} studies how different vision tasks are related to each other. \jiasen{Strezoski \etal \cite{strezoski2019many} studies layer-wise task routing for different vision tasks.}  
McCann \etal \cite{mccann2018natural} pose ten natural language processing (NLP) tasks as question answering tasks.
MT-DNN \cite{liu2019multi} combines multi-task learning with pretraining \cite{devlin2018bert} to improve the learning of text representations. 
Despite this progress, it is still challenging to train a single model on many tasks that can outperform or even match their single-task counterparts. To enhance the training scheme, BAM \cite{clark2019bam} applies knowledge distillation where single-task models teach the multi-task model.  
% about how to sample the dataset. 
Raffel \etal \cite{raffel2019exploring} explore different sampling strategies for NLP tasks. 
We focus on multi-task learning for V\&L tasks.% and automatic sampling strategies.

\xhdr{Vision and language.} 
% List old task specific define for vision and language model.
While we address 12 V\&L tasks in Sec.~\ref{sec:task_group}, we do miss some families of tasks including image and video captioning \cite{cococaption}, visual dialog \cite{visualdialog}, embodied question answering \cite{embodiedqa} and instruction following \cite{mattersim}.
% List recent pretrain-then-finetune approach. 
%
Different from earlier work \cite{lu2016hierarchical, yu2018mattnet, zhu2016visual7w, lu2018neural, hu2017modeling, zellers2019vcr, lee2018stacked} which design bespoke architecture for different tasks, 
recently proposed models for V\&L \cite{lu2019vilbert, tan2019lxmert, li2019visualbert, alberti2019fusion, li2019unicoder, su2019vl, zhou2019unified, chen2019uniter} provide a common architecture that can be pretrained using self-supervised losses and adapted to many vision and language tasks. However, these models still require task specific finetuining, which may easily overfit on small dataset.
% requires large number of parameters when number of tasks increased. 
Our single model jointly learns from multiple V\&L tasks and achieves competitive performance. Further, multi-task training provides a better visolinguistic representation for task specific finetuning than self-supervised objectives. 

\xhdr{Multi-task V\&L learning.} 
Recent work \cite{pramanik2019omninet, nguyen2019multi, shuster2019dialogue} also explores multi-task learning in V\&L. HDC \cite{nguyen2019multi} trains a multi-task network on multiple datasets and uses a hyper-parameter search method to determine which layer output should be taken for each task. Our method does not need any hyperparameter search to choose outputs for different tasks and outperforms both \cite{pramanik2019omninet} and  \cite{nguyen2019multi}. \cite{shuster2019dialogue} is a concurrent work that does multi-task training on 12 dialogue datasets (only two with images). Our work differs in that we focus on a variety of vision and language tasks.

\csection{Conclusion}

{In this work, we develop a training regime and experimental setting for large-scale, multi-modal, multi-task learning. As one part of this, we introduce a novel task scheduling approach to help avoid over- or under-training tasks with differing sizes or difficulties.
%Using this framework, we explore the relationships between vision-and-language datasets -- training a single model that performs at state of the art on 12 popular vision-and-language datasets.
{Using this framework, we explore the relationships between 12 vision-and-language datasets -- our single multi-task model outperforms 12 single-task models.
} %% Marcus: focus on single vs. multi-task first! sota is already below
We find multi-task training can lead to significant gains over independent task training. Further, we show that multi-task learning is an effective pre-training task for training state-of-the-art single-task models.}

{\footnotesize
\noindent \textbf{Acknowledgement.}
The GaTech effort was supported in part by NSF, AFRL, DARPA, ONR YIPs, ARO PECASE, Amazon. The views and conclusions contained herein are those of the authors and should not be interpreted as necessarily representing the official policies or endorsements, either expressed or implied, of the U.S. Government, or any sponsor.}

{\small
\bibliographystyle{style/ieee_fullname}
\bibliography{bib/main}
}

\clearpage

\twocolumn[{%
 \centering
 \LARGE 12-in-1: Multi-Task Vision and Language Representation Learning\\[1.5em]

}]

\csection{Supplementary}

{In this section, we first show the full details of the cleaned dataset in Sec.~\ref{sec:dataset}. We further discuss the modifications in pretraining, show our multi-task model architecture and describe the implementation details in  Sec.~\ref{sec:improve-vilbert}, Sec.~\ref{sec:suppliment_model_figure} and Sec.~\ref{sec:supp_implementation} respectively. The rest of the section provides extensive experimental results to fully analyze our proposed model.}

% \marcus{can we had a summary here pointing to the different subsections?}
% \marcus{Are the citation numbering still consistent with the paper we submitted as main paper?} Jiasen: I think so, we use the same bib to cite the paper. 
\medskip
\csubsection{Datasets}
\label{sec:dataset}
% Here we describe the datasets used and their details

Table~\ref{tab:datasets} shows the number of images in the train+val and test sets before and after cleaning. Our cleaning process removes 13.02\% of the total number of images on average. It is important to note that here we show the number of images per dataset and not the number of actual training samples. Different tasks have different number of training samples for each image. For details on training samples please refer to Table \ref{tab:training_details}. 
{We collect the union of all dataset test sets and remove any occurrence of these images from all training and validation sets; in this way we arrive at the \textit{Clean} training and validation sets. }{With this strategy, the test sets of the original datasets are not modified in any way.}

%We analyse the train and test sets for each task and find the pool of all test images. We remove any occurence of any of the images from this pool from all training sets of each task to arrive at the \textit{Clean} training set.

\begin{table}[ht]
\setlength\tabcolsep{3 pt}
\resizebox{\columnwidth}{!}{
  \begin{tabular}{@{}l@{}r r c | c  }
  \toprule
   & \multicolumn{1}{c}{Train+Val}  & \multicolumn{1}{c}{Test} & \multicolumn{1}{c}{Cleaned Train+Val}  & \multicolumn{1}{c}{\% Removed}  \\
  \midrule
 \texttt{[A]} VQA2.0\cite{goyal2017making} & 123,287 & 81,434 & 98,861 & 19.81  \\
 \texttt{[B]} VG QA\cite{krishna2017visual} & 108,249 & - & 92,147 & 14.87 \\
  \texttt{[C]} GQA\cite{hudson2019gqa} & 82,374 & 2,987 & 69,868 & 15.18  \\ 
 \texttt{[D]} COCO Retrieval\cite{cococaption} & 118,287 & 5,000 & 99,435 & 15.93  \\
 \texttt{[E]} Flickr30k Retrieval \cite{plummer2015flickr30k} & 30,014 & 1,000 & 29,077 & 3.12  \\
  \texttt{[F]} RefCOCO\cite{kazemzadeh2014referitgame} & 18,494 & 1,500 & 14,481 & 21.69  \\
  \texttt{[F]} RefCOCO+\cite{kazemzadeh2014referitgame} & 18,492 & 1,500 & 14,479 & 21.70   \\
  \texttt{[H]} RefCOCOG \cite{mao2016generation} & 23,199 & 2,600 & 17,903 & 22.82   \\
  \texttt{[I]} Visual 7W \cite{zhu2016visual7w} & 17,953 & 7,780 & 16,415 & 8.56   \\
   \texttt{[J]} GuessWhat\cite{de2017guesswhat} & 56,638 & 9,899 & 51,291 & 9.44  \\
   	\texttt{[K]} SNLI-VE\cite{xie2018visual} & 30,783 & 1,000 & 29,808 & 3.16   \\
    \texttt{[L]} NLVR$^2$ \cite{suhr2019corpus} & 95,522 & 8,056 & 95,522 & 0   \\
  \midrule
  Average & - & - & - & 13.02 \\
  \bottomrule
  \end{tabular}}
    \smallskip
  \caption{Number of images in the train+val and test sets before and after cleaning. We use the training part of the cleaned dataset in the multi-task experiments. Note that this is not the number of training samples but the number of images in the dataset.}
 \label{tab:datasets}
\end{table}

\csubsection{Improvements over ViLBERT Pretraining}
In this section, we discuss in detail the modification we made to the base ViLBERT pretraining approach. 
\label{sec:improve-vilbert}

\xhdr{Masked prediction with mislaigned pairs.} In the original ViLBERT pretraining procedure, the model observes an image and caption as inputs. The caption is either obtained from the paired caption (with $p=0.5$) or a randomly sampled misaligned caption from the dataset. The \textit{multi-modal alignment prediction} task, which predicts whether the image and caption are aligned, is crucial for image retrieval tasks \cite{lu2019vilbert, tan2019lxmert, li2019unicoder}. 
Recent work \cite{su2019vl} has questioned the necessity of the \textit{multi-modal alignment prediction} task and observed better performance on non-image retrieval tasks without this pretraining objective. Similar observations are also found in the natural language understanding tasks \cite{liu2019roberta, lample2019cross, yang2019xlnet, joshi2019spanbert}. 
% what we found the real reason:
Digging further into this, we find that both the alignment and prediction tasks are typically done together. For misaligned image-caption pairs, this amounts to forcing the model to predict missing image or text regions based on incorrect paired data! We find the model will learn worse context representations in this setup. 
% what we do:
Instead of removing the \textit{multi-modal alignment prediction} task, we only perform the \textit{mask multi-modal modelling} task on \textbf{aligned image-caption pairs}. This will effectively remove the noise introduced by negative samples.
% and we empirically show that this leads to better performances on all the downstream tasks. 

% Masking the regions. 
\xhdr{Masking overlapping regions.} Different from words embedding in the caption, visual feature embeddings (extracted from a pretrained Faster-RCNN \cite{ren2015faster}) have a lot of repetitions due to overlapped image regions. To avoid visual clue leakage from the visual embedding of other elements, VL-BERT \cite{su2019vl} sets the pixels laid in the masked RoI to zeros before applying Faster R-CNN. However, overlapped image patches with boundary information may still leak the visual clues for the masked RoI. 
% what we do here.
We mask the overlapped image regions in a more aggressive manner -- any visual embedding that overlaps a masked region by 40\% IOU or more is also masked. We observe significant improvements over the ViLBERT model as shown in Table~\ref{tab:finetuning} when comparing column ViLBERT with Ours$_\texttt{ST}$.

% \csubsection{Training details and hyperparameters}
\csubsection{Model Architecture}
\label{sec:suppliment_model_figure}

\begin{figure}[ht]
\centering
\includegraphics[width=\columnwidth]{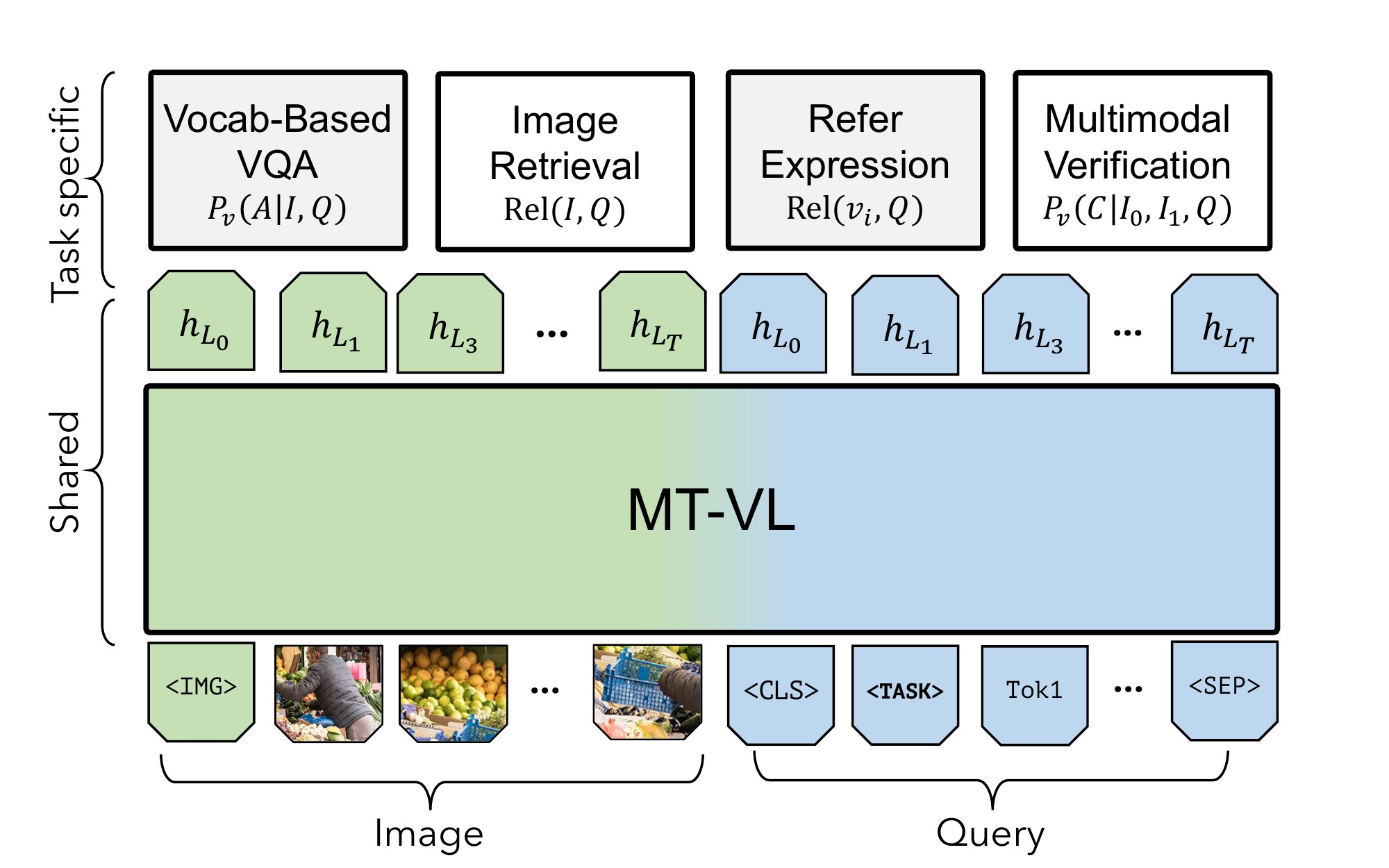}
\caption{Architecture of the our model for V\&L multi-task learning. We augment the input query with a task token to learn the task-aware feature embedding. }
%\vspace{-5 pt}
\label{fig:model_supp}
\end{figure}

Fig.~\ref{fig:model_supp} shows the architecture of the our model for V\&L multi-task learning, which is described in Sec.~\ref{sec:mt-vilbert}. We use ViLBERT as our base model shared across different tasks. For the task-specific heads, our model jointly train with four different task group -- Vocab-Based VQA; Image Retrieval, Refer Expression and Multimodal Verification. 

\csubsection{Implementation Details}
\label{sec:supp_implementation}

Image features are extracted from a ResNeXT-152 Faster-RCNN model trained on Visual Genome(VG) with attribute loss. \jiasen{Our model is first initialized from pretrained BERT weights \cite{devlin2018bert}. 
Our  models are trained using AdamW optimizer \cite{loshchilov2017fixing} with a linear warmup and linear decay learning rate scheduler. We train our multi-task model for ~40K total iterations (same as the number of iterations for the VG QA single task) on 8 NVIDIA V100 GPUs for 5 days. We use AdamW optimizer and a warmup linear schedule. Hyperparameters like learning rate and batch sizes used for each task are listed in Table~\ref{tab:training_details}. We also report the number of training samples used in various settings in our experiments.}

\begin{table}[t]
\setlength\tabcolsep{3 pt}
\resizebox{\columnwidth}{!}{
  \begin{tabular}{@{}l@{} rrr | c | r  c@{}}
  \toprule
     & \multicolumn{3}{c}{Samples} & \multicolumn{1}{c}{}  & \multicolumn{2}{c}{Hyperparams}  \\ 
   & \multicolumn{1}{c}{Full Train}  & \multicolumn{1}{c}{Cleaned Train} & \multicolumn{1}{c}{Test} & \multicolumn{1}{c}{Metric}  & \multicolumn{1}{c}{BS} & \multicolumn{1}{c}{LR}  \\ 
  \midrule
 \texttt{[A]} VQA2.0\cite{goyal2017making} & 655,111 & 542,104 & 447,793 & VQA Accuracy & 128 & 4e-5  \\
 \texttt{[B]} VG QA\cite{krishna2017visual} & 1,437,931 & 1,294,255 & 5,000 & VQA Accuracy & 128 & 4e-5  \\
  \texttt{[C]} GQA\cite{hudson2019gqa} & 1,072,062 & 962,928 & 12,578 & VQA Accuracy & 128 & 4e-5 \\ 
 \texttt{[D]} IR COCO \cite{cococaption} & 566,747 & 487,600 & 1,000 & Recall @ 1, 5, 10 & 128 & 2e-5  \\
 \texttt{[E]} IR Flickr30k \cite{plummer2015flickr30k} & 145,000 & 140,485 & 1,000 & Recall @ 1, 5, 10 & 128 & 2e-5 \\
  \texttt{[F]} RefCOCO\cite{kazemzadeh2014referitgame} & 120,624 & 96,221 & 10,752 & Accuracy & 256 & 2e-5  \\
  \texttt{[F]} RefCOCO+\cite{kazemzadeh2014referitgame} & 120,191 & 95,852 & 10,615 & Accuracy & 256 & 2e-5   \\
  \texttt{[H]} RefCOCOG \cite{mao2016generation} & 80,512 & 65,514 & 9,602 & Accuracy & 256 & 2e-5   \\
  \texttt{[I]} Visual 7W \cite{zhu2016visual7w} & 93,813 & 93,813 & 57,265 & Accuracy & 256 & 2e-5  \\
   \texttt{[J]} GuessWhat\cite{de2017guesswhat} & 113,221 & 100,398 & 23,785 & Accuracy & 64 & 2e-5  \\
   	\texttt{[K]} NLVR$^2$ \cite{suhr2019corpus} & 86,373 & 86,373 & 6,967 & Accuracy & 64 & 2e-5   \\
    \texttt{[L]} SNLI-VE\cite{xie2018visual} & 529,527 & 512,396 & 17,901 & Accuracy & 256 & 2e-5   \\
  \midrule
  Total & 5,021,112 & 4,477,939 & 604,258 & - & - & -  \\
  \bottomrule
  \end{tabular}}
    \smallskip
  \caption{Training details including sample sizes, testing metric and hyperparameters for single task and multi-task training.}
 \label{tab:training_details}
\end{table}

\csubsection{Multi-Task Training}
\label{sec:supp_mt_training}

\begin{figure*}[ht]
\centering
\includegraphics[width=0.99\textwidth]{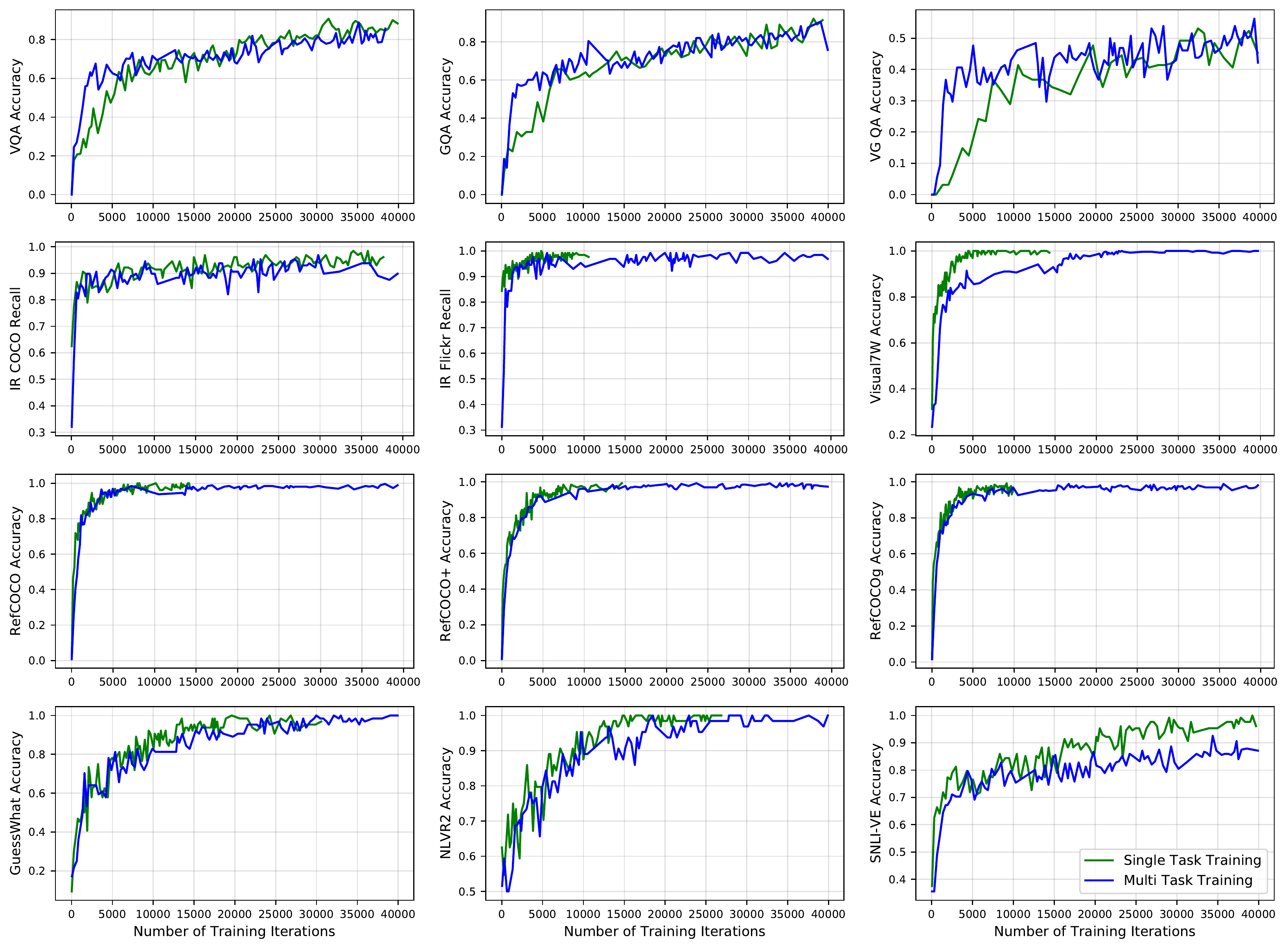}
% \vspace{3pt}
\caption{Training curves on \textit{train} set for Ours$_\texttt{ST}$ (Table~\ref{tab:full} \texttt{Row 2}) vs Ours$_\texttt{AT}$ (Table~\ref{tab:full} \texttt{Row 4}) models for all the 12 tasks in our experiments. Green lines show single-task training(Ours$_\texttt{ST}$) and blue lines show multi-task training(Ours$_\texttt{AT}$). {Note that all these training are with the \textit{Clean V\&L} setup. We can observe that for some of the tasks the training for Ours$_\texttt{ST}$ are shorter as they have fewer number of iterations when trained alone. Please refer to Sec.~\ref{sec:supp_mt_training} for more details.}
% \marcus{What is the message from them?}\marcus{Is it on the training or validation set?} \marcus{Add a comment why Single Task Training stops earlier} \marcus{What is the message from them?} 
%\marcus{Does this correspond to the training runs in line 2 and 4 in Table 2? Then we should add that. Also let's use the same naming, as we have several ST and several multi-task models.}
}
\vspace{-5 pt}
\label{fig:training_all}
\end{figure*}

To further illustrate the multi-task training process, in Fig. \ref{fig:training_all} we show the training curves for single-task \vs multi-task for all the 12 tasks in our setup. 
Green lines show single-task training and blue lines show multi-task training. Since we train the model with maximum iterations across different datasets for multi-task training, for some smaller datasets (\eg RefCOCO, Visual7W \etc), the number of iterations for single task is much smaller compared to the multi-task setting. By comparing the training curves of single-tasks and multi-tasks, we can see that most of the tasks have similar training curves. However, the tasks in the vocab-based VQA group benefit from the multi-task training with faster convergence within first 10000 iterations.

\begin{figure}[t]
\centering
\includegraphics[width=\columnwidth]{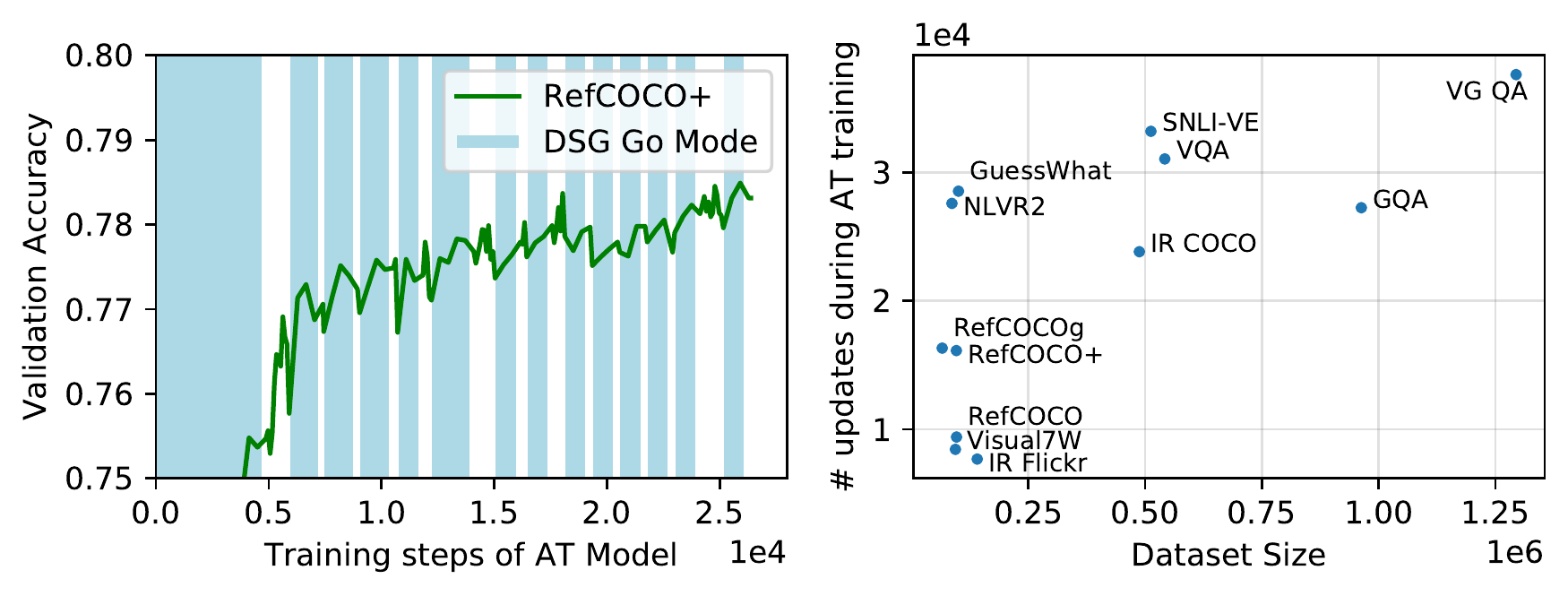}
\caption{Left: Val acc.~of our AT model on RefCOCO+. Right: Dataset size vs.~number of updates during stop-and-go training.}
\label{fig:rebutal}
\end{figure}

% \marcus{What is the message from them?}

\jiasen{\xhdr{Concept drift of smaller datasets.}
In Fig.~\ref{fig:rebutal}~(left), we plot the val accuracy of our AT model on RefCOCO+ to show the concept drift of smaller datasets during MT training. Even with sparse updates (stop mode), we observe sharp drops (dips before go mode is reactivated)
on RefCOCO+. 

\xhdr{Relationship between dataset size and \texttt{go} mode duration.} The dataset size gap can be significant -- up to 16:1 for VG QA vs ReferCOCOg. To illustrate how dataset size affects our dynamic stop-and-go training regime, we plot dataset size vs active training iterations in Fig.~\ref{fig:rebutal} (right). Among datasets with a similar size, we see significant differences in active training time. This shows that dynamic stop-and-go addresses issues of dataset difficulties rather than just size. However, there is a general trend that larger datasets do tend to stay in the active \texttt{go} mode longer. 

\xhdr{Full per task accuracy for  $\texttt{AT}$ without G4 model.} In Table~\ref{tab:full}, we observed from the representative task analysis that G4 tends to have a negatively effect other group during joint training. In Table~\ref{tab:ablation_nog4_appendix}, we further show full per task accuracy for $\texttt{AT}$ without G4 model and different ablations. We can see that $\texttt{AT}_{\mbox{\small w/o G4}}$ outperforms $\texttt{AT}$ 0.48\% on MT scores, which verifies G4 tends to have negatively effect even on the finetuned model. How to remove the negative interactions between different tasks is left to future study. }

\csubsection{Comparison with other SOTA}
\label{sec:comparsion}

Table~\ref{tab:finetuning} shows the detailed comparison of Ours$_\texttt{ST}$ {(also shown in Table \ref{tab:full}, line 1)} and Ours$_\texttt{AT->ST}$ {(also shown in Table \ref{tab:full}, line 8)} with the recent SOTA approaches, inlcuding ViLBERT \cite{lu2019vilbert}, Unicoder-VL \cite{li2019unicoder}, VisualBERT \cite{li2019visualbert}, LXMERT \cite{tan2019lxmert} and UNITER \cite{chen2019uniter}. Most of the recent proposed methods follows the pretrain-then-finetune scheme, usually pretraining on out-of-domain data or in-domain data. The out-of-domain data contains Conceptual Caption Dataset (CC) \cite{sharma2018conceptual} and SBU dataset \cite{ordonez2011im2text} while in-domain data contains COCO \cite{cococaption} and Visual Genome \cite{krishnavisualgenome}.  Pre-training on the in-domain datasets usually leads to better downstream performance, since there is less domain transfer from pretraining to finetuning. Similar to ViLBERT, we pretrain our model on CC, which is different from VLBERT (CC + Wiki Corpus), VisualBERT (CC + COCO), LXMERT (COCO + VG) and UNITER (CC + SUB + COCO + VG). We achieve comparable performance with less pretrained data.

\begin{table*}[ht]\footnotesize
\setlength\tabcolsep{5 pt}
\centering
\resizebox{\textwidth}{!}{
 \begin{tabular}{l | c | c | c | c | c | c | c | c c | c  c}
 \toprule
\multirow{2}{*}{Tasks} &  & \multirow{2}{*}{\footnotesize{SOTA}} & \multirow{2}{*}{\footnotesize{ViLBERT}} & \multirow{2}{*}{\footnotesize{VLBERT}}   & \multirow{2}{*}{\footnotesize{Unicoder-VL}} & \multirow{2}{*}{\footnotesize{VisualBERT}} &\multirow{2}{*}{\footnotesize{LXMERT}} & \multicolumn{2}{c}{\footnotesize{UNITER}} & \multirow{2}{*}{\footnotesize{Ours$_\texttt{ST}$}} & \multirow{2}{*}{\footnotesize{Ours$_\texttt{AT->ST}$}} \\
& & & & & & & & \footnotesize{BASE} & \footnotesize{LARGE} & &  \\
\midrule
 & Pretraining Data & & \footnotesize{CC} & \footnotesize{CC + Wiki Corpus}  & \footnotesize{CC} & \footnotesize{CC + COCO} & \footnotesize{COCO + VG} & \multicolumn{2}{c}{\footnotesize{CC+SUB+COCO+VG}} & \footnotesize{CC} & \footnotesize{CC}\\
\midrule
\multirow{1}{*}{\small{VQA}} & test-dev & 70.63 & 70.55& 70.50& - & 70.80 &  72.42 & 72.27 & \textbf{73.24} & 71.82 & 73.15 \\
%  & test-std & 70.90  & 70.92 & 70.83 & - & 71.00 &  72.54 & 72.46 & \textbf{73.40}  & - & - \\
  \midrule
\small{VG QA} & val & - & - & - & - & - &  - & - & -  & 34.38 &  \textbf{36.64} \\
  \midrule
\multirow{1}{*}{\small{GQA}} & test-dev & - & - & - & - & - & 60.00 & - & - & 58.19 & \textbf{60.65 } \\
%  & test-std &   &  &  & - & &   & &   & - & - & - \\
  \midrule
%  \multirow{3}{*}{\small{Retrieval COCO}} & R1 & 38.60 & -& -& 44.50 & - & - & 48.42& \textbf{51.72} & & \\
%  & R5 & 69.30 & - & - & 74.40&-& -& 76.68 & \textbf{78.41} & & \\
%   & R10 & 80.40 & - & - & 84.00 &-& -& 85.90& \textbf{86.93} & & \\
 \multirow{3}{*}{\small{IR COCO}} & R1 &  61.60 & -& -& \textbf{68.50} & - & - & - & - & 65.28 & 68.00 \\
 & R5 & 89.6 & - & - & \textbf{92.70} &-& -& - & - & 91.02 & 92.38 \\
  & R10 & 95.2 & - & - & \textbf{96.90} &-& -& - & - & 96.18 & 96.52 \\
  \midrule
 \multirow{3}{*}{\small{IR Flickr}} & R1 & 48.60 & 58.20& -& 68.30 & - & - &  71.50 & \textbf{73.66} & 61.14 & 67.90 \\
 & R5 & 77.70 & 84.90 & - & 90.30&-& -&  91.16 & \textbf{93.06} & 87.16 & 89.60 \\
  & R10 & 85.20 & 91.52 & - & 94.60 &-& -&  95.20 & \textbf{95.98} & 92.30 & 94.18 \\
  \midrule
\small{Visual 7W} & test & 72.53 & - & - & - & - & - & - & - & 80.51 & \textbf{83.35} \\
  \midrule
 \small{Ref-COCO} & test & 77.12 & - & - & -&-& - & 80.48 & 80.88  & 78.63 & \textbf{81.20} \\
  \midrule
 \small{Ref-COCO+}& test & 67.17 & 70.93 & 69.47 & - & - & - & 73.26 & 73.73  & 71.11 & \textbf{74.22} \\
  \midrule
\small{Ref-COCOg} & test & 69.46 & - & - & - & - & - & 74.51 & 75.77  & 72.24 & \textbf{76.35} \\
  \midrule
\small{GuessWhat} & test & 61.30 & - & - & - &-& -& - & -  & 62.81 & \textbf{65.69} \\
  \midrule
 
\small{NLVR$^2$} & test-P & 53.50 & - & - & - & 67.00 & 74.50 & 77.87 & \textbf{79.50}  & 74.25 & 78.87 \\
  \midrule
\small{SNLI-VE} & test & 71.16 & - & - & - & - & - & 78.02 & \textbf{78.98}  & 76.72 & 76.95\\
\bottomrule
\end{tabular}}
\smallskip
\caption{Comparison of Ours$_\texttt{ST}$ (Table~\ref{tab:full} \texttt{Row 1}) and Ours$_\texttt{AT->ST}$ (Table~\ref{tab:full} \texttt{Row 8}) models on full dataset with other SOTA methods. Results for RefCOCO and RefCOCO+ are reported on the full test split (testA + testB). Refer to Sec~\ref{sec:comparsion} for more details.
% \marcus{Mention the important observations from the text here. E.g. Vilbert vs. Ours} \vedanuj{This is discussed in the Sec 8.6}
% \marcus{Add a row number of params as in Table 4?} \vedanuj{Number of params doesn't make much sense IMO here as we are not showing any multi-task model}
}
\label{tab:finetuning}
\end{table*}

\csubsection{Full Breakdown of Ablation Study}
\label{sec:full_ablation}
Table~\ref{tab:ablation_appendix} shows the full breakdown of Table~\ref{tab:ablation} and 
Fig.~\ref{fig:iter_gap} per task in the main paper. RC refers to Retrieval COCO and RF refers to Retrieval Flickr30k. VQA and GQA are evaluated on \texttt{test-dev} splits. Retrieval COCO and Flickr30k are evaluated on their respective 1K test split. NLVR$^2$ is evaluated on \texttt{testP} split. All other datasets are evaluated on their respective test splits. Table~\ref{tab:ablation_dsg_appendix} shows the full scores for each task for different \texttt{DSG} iteration gap ($\Delta$). Table~\ref{tab:ablation_nog4_appendix} shows the detailed per task scores for $\texttt{AT}_{\mbox{\small w/o G4}}$ model and different ablations for it. We compare with full \texttt{AT} model as well. 

\begin{table*}[ht]\footnotesize
\centering
\fontsize{7pt}{7pt}\selectfont
\newcolumntype{C}{>{\centering\arraybackslash}X}
\setlength{\tabcolsep}{0pt}
\setlength{\extrarowheight}{5pt}
\renewcommand{\arraystretch}{0.75}
\begin{tabularx}{\textwidth}{p{10pt}p{2cm}!{\color{lightgray}\vline} CCCC!{\color{lightgray}\vline}CCCCCCC!{\color{lightgray}\vline}CCCCCC!{\color{lightgray}\vline}CCC!{\color{lightgray}\vline}C}
\toprule
 &
 & \rotatebox{90}{\raisebox{0.5pt} VQA}
 & \rotatebox{90}{\raisebox{0.5pt} VG QA}
 & \rotatebox{90}{\raisebox{0.5pt} GQA}
 & \rotatebox{90}{\raisebox{0.5pt} Mean G1}
 & \rotatebox{90}{\raisebox{0.5pt} RC R@1}
 & \rotatebox{90}{\raisebox{0.5pt} RC R@5}
 & \rotatebox{90}{\raisebox{0.5pt} RC R@10}
 & \rotatebox{90}{\raisebox{0.5pt} RF R@1}
 & \rotatebox{90}{\raisebox{0.5pt} RF R@5}
 & \rotatebox{90}{\raisebox{0.5pt} RF R@10}
 & \rotatebox{90}{\raisebox{0.5pt} Mean G2 (R1)}
 & \rotatebox{90}{\raisebox{0.5pt} RefCOCO}
 & \rotatebox{90}{\raisebox{0.5pt} RefCOCO+}
 & \rotatebox{90}{\raisebox{0.5pt} RefCOCOG}
 & \rotatebox{90}{\raisebox{0.5pt} Visual 7W}
 & \rotatebox{90}{\raisebox{0.5pt} GuessWhat}
 & \rotatebox{90}{\raisebox{0.5pt} Mean G3}
 & \rotatebox{90}{\raisebox{0.5pt} NLVR$^2$}
 & \rotatebox{90}{\raisebox{0.5pt} SNLI-VE}
 & \rotatebox{90}{\raisebox{0.5pt} Mean G4}
 & \rotatebox{90}{\raisebox{0.5pt} MT Score}\\
 
\midrule
& token per dataset & \bf 72.57 & \bf 36.36 & 60.12 & \bf 56.35 & \bf 63.70 & \bf 90.84 & \bf 96.16 & \bf 63.52 & \bf 87.48	& \bf 93.16	& \bf 63.61	& 80.58	& 73.25	& \bf 75.96	& 82.75	& \bf 65.04	& \bf 75.52	& \bf 78.44	& 76.78 & \bf 77.61	& \bf 69.08	 \\
 & token per head & 72.11	& 35.84	& 59.91	& 55.95	& 60.66	& 88.96	& 94.86	& 62.30	& 86.20	& 92.00	& 61.48	& \bf 80.67	& 73.10	& 75.82	& \bf 82.92	& 64.24	& 75.35	& 77.65	& \bf 77.08	& 77.37	& 68.52	\\
& w/o task token & 72.00 & 35.09 & 59.92	& 55.67	& 63.16	& 90.48	& 95.44	& 61.94	& 86.96	& 92.88	& 62.55	& 80.32	& 73.04	& 75.94	& 82.72	& 64.89	& 75.38	& 76.99	& 76.46	& 76.73	& 68.53 \\
& w/o \texttt{DSG} & 71.99	& 35.59	& 58.93	& 55.50	& 62.54	& 90.08	& 95.42	& 63.30	& 86.98	& 92.86	& 62.92	& 79.99	& 73.09	& 75.94	& 82.68	& 64.52	& 75.24	& 77.37	& 76.31	& 76.84	& 68.52 \\

\arrayrulecolor{lightgray}\specialrule{.5pt}{0.6pt}{-0.5pt}\arrayrulecolor{black}

& w/ curriculum & 70.59	& 35.54	& 57.91	& 54.68	& 61.14	& 89.74	& 95.04	& 61.28	& 86.58	& 92.56	& 61.21	&	80.11	& 73.35	& 75.62	& 82.38	& 64.51	& 75.19	& 77.20 & 76.19	& 76.69	& 67.98 \\
& w/ anti-curriculum & 71.53	& 35.54	&	\bf 60.39	&	55.82	&	61.04	& 88.78	& 94.96	&	58.12	&	84.66	&	90.84	&	59.58	&	78.99	&	71.34	&	74.24	&	80.80	&	63.08	&	73.69	&	76.14	&	75.74	&	75.94	&	67.24 \\

\arrayrulecolor{lightgray}\specialrule{.5pt}{0.6pt}{-0.5pt}\arrayrulecolor{black}
& vanilla multitask & 70.39	& 33.31	& 58.57	&	54.09	&	61.50	& 89.72	& 95.42	&	61.40	&	87.04	&	92.74	&	61.45	&	80.42 &	\bf 73.51 &	75.53 &	82.48 &	64.50 &	75.28 &	77.09 &	76.34 &	76.71	&	67.92 \\
\arrayrulecolor{lightgray}\specialrule{.5pt}{0.6pt}{-0.5pt}\arrayrulecolor{black}
& w/o CC pretraining & 70.23	& 33.49	& 58.41	&	54.04	&	57.92	& 87.60	& 93.96	&	56.72	&	83.20	&	90.68	&	57.32	&	77.93 &	69.60 &	72.21 &	78.99 &	61.67 &	72.08 &	73.63 &	75.92 &	74.77	&	65.56 \\
\bottomrule
\end{tabularx}
\smallskip
\caption{Full per task accuracy for the different ablation studies (summarized in Table~\ref{tab:ablation}). RC is Retrieval COCO and RF is Retrieval Flickr30k. Mean of G2 is taken over the Recall@1 scores. {We can see that with task token per dataset and \texttt{DSG} achieve the best performance.}}
\label{tab:ablation_appendix}
\end{table*}

\begin{table*}[ht]\footnotesize
\centering
\fontsize{7pt}{7pt}\selectfont
\newcolumntype{C}{>{\centering\arraybackslash}X}
\setlength{\tabcolsep}{0pt}
\setlength{\extrarowheight}{5pt}
\renewcommand{\arraystretch}{0.75}
\begin{tabularx}{\textwidth}{p{10pt}p{2cm}!{\color{lightgray}\vline} CCCC!{\color{lightgray}\vline}CCCCCCC!{\color{lightgray}\vline}CCCCCC!{\color{lightgray}\vline}CCC!{\color{lightgray}\vline}C}
\toprule
 &
 & \rotatebox{90}{\raisebox{0.5pt} VQA}
 & \rotatebox{90}{\raisebox{0.5pt} VG QA}
 & \rotatebox{90}{\raisebox{0.5pt} GQA}
 & \rotatebox{90}{\raisebox{0.5pt} Mean G1}
 & \rotatebox{90}{\raisebox{0.5pt} RC R@1}
 & \rotatebox{90}{\raisebox{0.5pt} RC R@5}
 & \rotatebox{90}{\raisebox{0.5pt} RC R@10}
 & \rotatebox{90}{\raisebox{0.5pt} RF R@1}
 & \rotatebox{90}{\raisebox{0.5pt} RF R@5}
 & \rotatebox{90}{\raisebox{0.5pt} RF R@10}
 & \rotatebox{90}{\raisebox{0.5pt} Mean G2 (R1)}
 & \rotatebox{90}{\raisebox{0.5pt} RefCOCO}
 & \rotatebox{90}{\raisebox{0.5pt} RefCOCO+}
 & \rotatebox{90}{\raisebox{0.5pt} RefCOCOG}
 & \rotatebox{90}{\raisebox{0.5pt} Visual 7W}
 & \rotatebox{90}{\raisebox{0.5pt} GuessWhat}
 & \rotatebox{90}{\raisebox{0.5pt} Mean G3}
 & \rotatebox{90}{\raisebox{0.5pt} NLVR$^2$}
 & \rotatebox{90}{\raisebox{0.5pt} SNLI-VE}
 & \rotatebox{90}{\raisebox{0.5pt} Mean G4}
 & \rotatebox{90}{\raisebox{0.5pt} MT Score}\\
 
\midrule
& \texttt{DSG} $\Delta 1$ & 71.99	& 35.59	& 58.93	& 55.50	& 62.54	& 90.08	& 95.42	& 63.30	& 86.98	& 92.86	& 62.92	& 79.99	& 73.09	& 75.94	& 82.68	& 64.52	& 75.24	& 77.37	& 76.31	& 76.84	& 68.52 \\
& \texttt{DSG} $\Delta 4$ & 72.57 & 36.36 & \textbf{60.12} & \textbf{56.35} & 63.70 & 90.84 & \textbf{96.16} & 63.52 & 87.48	& \textbf{93.16}	& 63.61	& 80.58	& 73.25	& \textbf{75.96}	& 82.75	&  65.04	& 75.52	& \textbf{78.44}	& \textbf{76.78} &\textbf{ 77.61}	& 69.08	 \\
& \texttt{DSG} $\Delta 8$  & 72.61 &	\textbf{36.65}	&	59.69 &	56.32	&	\textbf{65.24}	&	90.86	&	96.02	&	\textbf{63.56}	&	87.60	&	93.08	&	\textbf{64.40}	&	80.32	&	\textbf{73.56}	&	75.88	&	\textbf{82.79}	&	\textbf{65.33}	&	\textbf{75.58}	&	77.43	&	76.75	&	77.09	&	\textbf{69.15} \\
& \texttt{DSG} $\Delta 16$  & \textbf{72.74}	&	35.34	&	59.70	&	55.93	&	64.78	&	\textbf{91.04}	&	95.86	&	62.36	&	\textbf{87.66}	&	92.92	&	63.57	&	\textbf{80.59}	&	73.17	&	75.88	&	82.61	&	64.79	&	75.41	&	78.18	&	76.66	&	77.42	&	68.90 \\
\bottomrule
\end{tabularx}
\smallskip
\caption{Full per task accuracy for Fig.~\ref{fig:iter_gap} showing different Dynamic Stop-and-Go Iteration Gaps ($\Delta$). Mean of G2 is taken over the Recall@1 scores.}
\label{tab:ablation_dsg_appendix}
\end{table*}

\begin{table*}[ht]\footnotesize
\centering
\fontsize{7pt}{7pt}\selectfont
\newcolumntype{C}{>{\centering\arraybackslash}X}
\setlength{\tabcolsep}{0pt}
\setlength{\extrarowheight}{5pt}
\renewcommand{\arraystretch}{0.75}
\begin{tabularx}{\textwidth}{p{10pt}p{2cm}!{\color{lightgray}\vline} CCCC!{\color{lightgray}\vline}CCCCCCC!{\color{lightgray}\vline}CCCCCC!{\color{lightgray}\vline}C}
\toprule
 &
 & \rotatebox{90}{\raisebox{0.5pt} VQA}
 & \rotatebox{90}{\raisebox{0.5pt} VG QA}
 & \rotatebox{90}{\raisebox{0.5pt} GQA}
 & \rotatebox{90}{\raisebox{0.5pt} Mean G1}
 & \rotatebox{90}{\raisebox{0.5pt} RC R@1}
 & \rotatebox{90}{\raisebox{0.5pt} RC R@5}
 & \rotatebox{90}{\raisebox{0.5pt} RC R@10}
 & \rotatebox{90}{\raisebox{0.5pt} RF R@1}
 & \rotatebox{90}{\raisebox{0.5pt} RF R@5}
 & \rotatebox{90}{\raisebox{0.5pt} RF R@10}
 & \rotatebox{90}{\raisebox{0.5pt} Mean G2 (R1)}
 & \rotatebox{90}{\raisebox{0.5pt} RefCOCO}
 & \rotatebox{90}{\raisebox{0.5pt} RefCOCO+}
 & \rotatebox{90}{\raisebox{0.5pt} RefCOCOG}
 & \rotatebox{90}{\raisebox{0.5pt} Visual 7W}
 & \rotatebox{90}{\raisebox{0.5pt} GuessWhat}
 & \rotatebox{90}{\raisebox{0.5pt} Mean G3}
 & \rotatebox{90}{\raisebox{0.5pt} MT Score}\\
 
\midrule

& \texttt{AT} & 72.57 & 36.36 & 60.12 & 56.35 & 63.70 & 90.84 & 96.16 & 63.52 & 87.48	& 93.16	& 63.61	& 80.58	& 73.25	& \textbf{75.96}	& 82.75	&  65.04	& 75.52	& 56.15	 \\
\arrayrulecolor{lightgray}\specialrule{.5pt}{0.6pt}{-0.5pt}\arrayrulecolor{black}
& $\texttt{AT}_{\mbox{w/o G4}}$ & \bf 72.68 & \bf 36.74 & \bf 62.09 & \bf 57.17 & \bf 64.88 & \bf 91.36 & 95.98  &  \bf 64.62 & 87.98 & 93.18 & \bf 64.75 & \bf 80.76 & \bf 73.60 & 75.80 & \bf 83.03 & \bf 65.41 & \bf 75.72	& \bf 56.63 \\

& w/o task token  & 71.54 & 34.42 & 61.62 & 55.86 & 64.34 & 90.80 & \bf 96.18 & 63.24 & 86.86 & 92.52 & 63.79 & 80.53 & 72.77 & 75.33 & 82.79 & 64.52 & 75.18	& 55.92 \\
& w/o \texttt{DSG}  & 71.70 & 34.15 & 59.82 & 55.22 & 63.20 & 90.70 & 96.04 & 63.44 & \bf 88.18 & \bf 93.28 & 63.32 & 80.64 & 72.86 & 75.81 & 82.56 & 64.76 & 75.32	&	55.74 \\
\bottomrule
\end{tabularx}
\smallskip
\caption{Full per task accuracy for $\texttt{AT}_{\mbox{\small w/o G4}}$ model and different ablations for $\texttt{AT}_{\mbox{\small w/o G4}}$. Mean of G2 is taken over the Recall@1 scores.}
\label{tab:ablation_nog4_appendix}
\end{table*}

\csubsection{Multi-task visual grounding consistency}

In Sec.~\ref{sec:ablation_analysis}, we propose the multi-task visual grounding consistency. Here, we explain the proposed metric in more detail. 
Given $N$ images with RefCOCO/+ refer expression and VQA questions, we want to test if multi-task models exhibit more consistent visual groundings than independent task-specific models. For each image $I_i$, there are associated VQA question $\{q^{(i)}\}$ and referring expression $\{r^{(i)} \}$. To measure the overlap in visual concepts between a question $q_j^{(i)}$ and reference $r_k^{(i)}$, we count the the number of overlapped noun / adj as $d(q_j^{(i)}, r_k^{(i)})$, the multi-task visual grounding consistency can be calculated as:  
% Equation here:
\begin{equation}
\texttt{MT-VGC} = \frac{\sum_{k=0}^N \lvert \sum_{j} \sum_{k} d(q_j^{(i)},r_k^{(i)}) \mathbbm{1}_{\{y(q_j^{(i)}) = 1 \& y(r_k^{(i)}) = 1 \}} \rvert}{ \sum_{i=0}^N \lvert \sum_{j} \sum_{k} d(q_j^{(i)},r_k^{(i)}) \mathbbm{1} \rvert}
\end{equation}
where $y(q_k^{(i)}) = 1$ means the model correctly answer the question $q_k^{(i)}$ based on VQA accuracy metric and $y(r_k^{(i)}) = 1$ means the model correctly locate the image regions (IoU > 0.5) given the reference $r_k^{(i)}$.

%In \todo{Fig}.~\ref{}, we show the distributions of the overlapping counts across 111,275 pairs from 1500 images.

\csubsection{Qualitative Results}
\label{sec:qualitative}

\begin{figure*}[t]
\centering
\fbox{\includegraphics[width=0.99\textwidth]{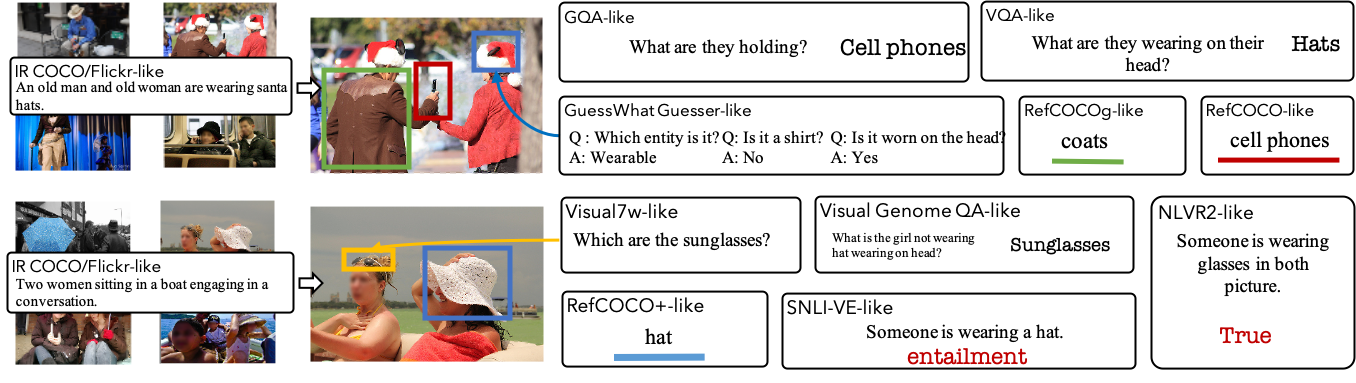}}
\fbox{\includegraphics[width=0.99\textwidth]{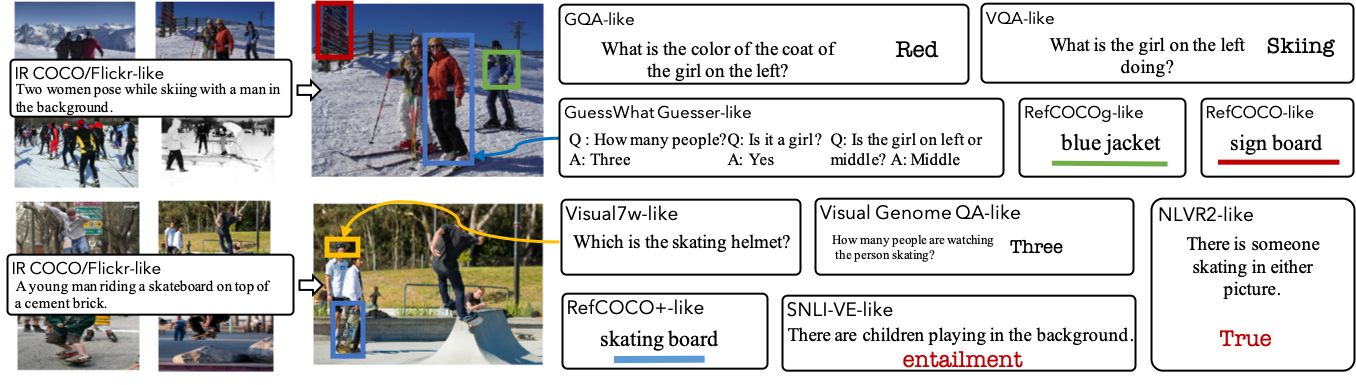}}
\vspace{3pt}
\caption{{\textbf{Our single multi-task model can solve multiple task consistently and correctly.}} Additional qualitative examples of our single model Our$_\texttt{AT}$ on multitude of V\&L tasks: caption and image retrieval, question answering, grounding phrases, guessing image regions based on a dialog, verifying facts about a pair of images, natural language inferences from an image, etc. Here we show outputs of our model for a variety of inputs (that mimic tasks from the 12 datasets it has been trained on). }
\vspace{-10 pt}
\label{fig:qual2}
\end{figure*}

\begin{figure*}[ht]
\centering
\includegraphics[width=\textwidth]{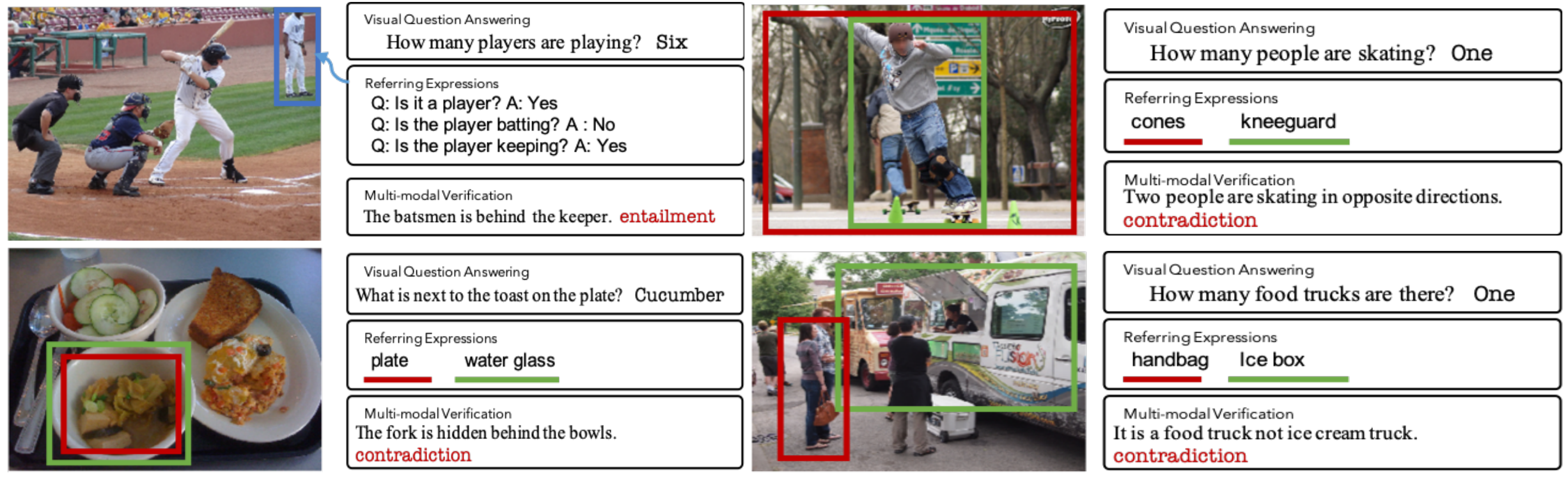}
\caption{Failure cases of our single \texttt{AT} model on multitude of V\&L tasks. Failure cases mostly occur when the model encounters \texttt{counting} questions or difficult referring expressions and phrases for fine grained recognition. 
}
% \marcus{would be good to add at least 3 examples}
\vspace{-8 pt}
\label{fig:failure}
\end{figure*}

Fig.~\ref{fig:qual2} shows more qualitative examples of our single model Our$_\texttt{AT}$ on different vision and language tasks and Fig.~\ref{fig:failure} shows some failure cases.
% \marcus{Any insights you got from qualitative looking at examples? Any observations you made? What are typical kind of failures? } 
The examples in Fig.~\ref{fig:qual2} show that the \texttt{AT} model works well for these wide range of tasks consistently. It can perform well in both short as well as long reasoning questions, image retrieval, pointing tasks, referring expressions and multi-modal validation. Failure cases mostly occur when the model encounters \texttt{counting} questions or difficult referring expressions and phrases for fine grained recognition.

\begin{figure*}[t]
\centering
\includegraphics[width=0.99\textwidth]{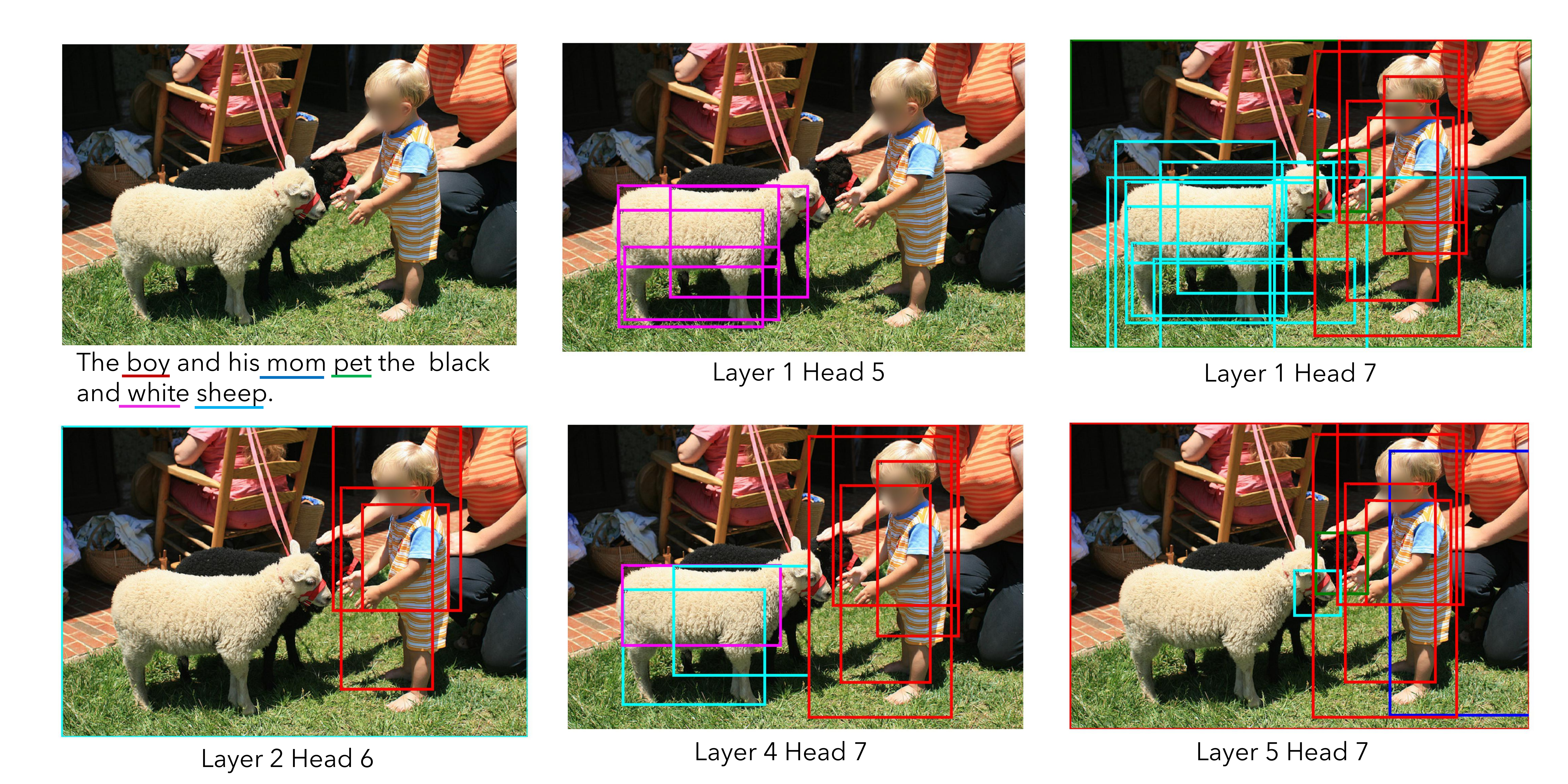}
\vspace{3pt}
\caption{Visualizations of image to sentence attention for the pretrained model on conceptual caption dataset. Given the image to sentence co-attention, we use the most attended word to represent its semantic meaning, and show the patches corresponding to the visual words (`boy', `mom', `pet', `white', `sheep'). Different colors show a correspondence between attended regions and underlined words. We can see that the model learns meaningful concept through pretraining.}
\vspace{-10 pt}
\label{fig:attention_0}
\end{figure*}

\begin{figure*}[t]
\centering
\includegraphics[width=0.99\textwidth]{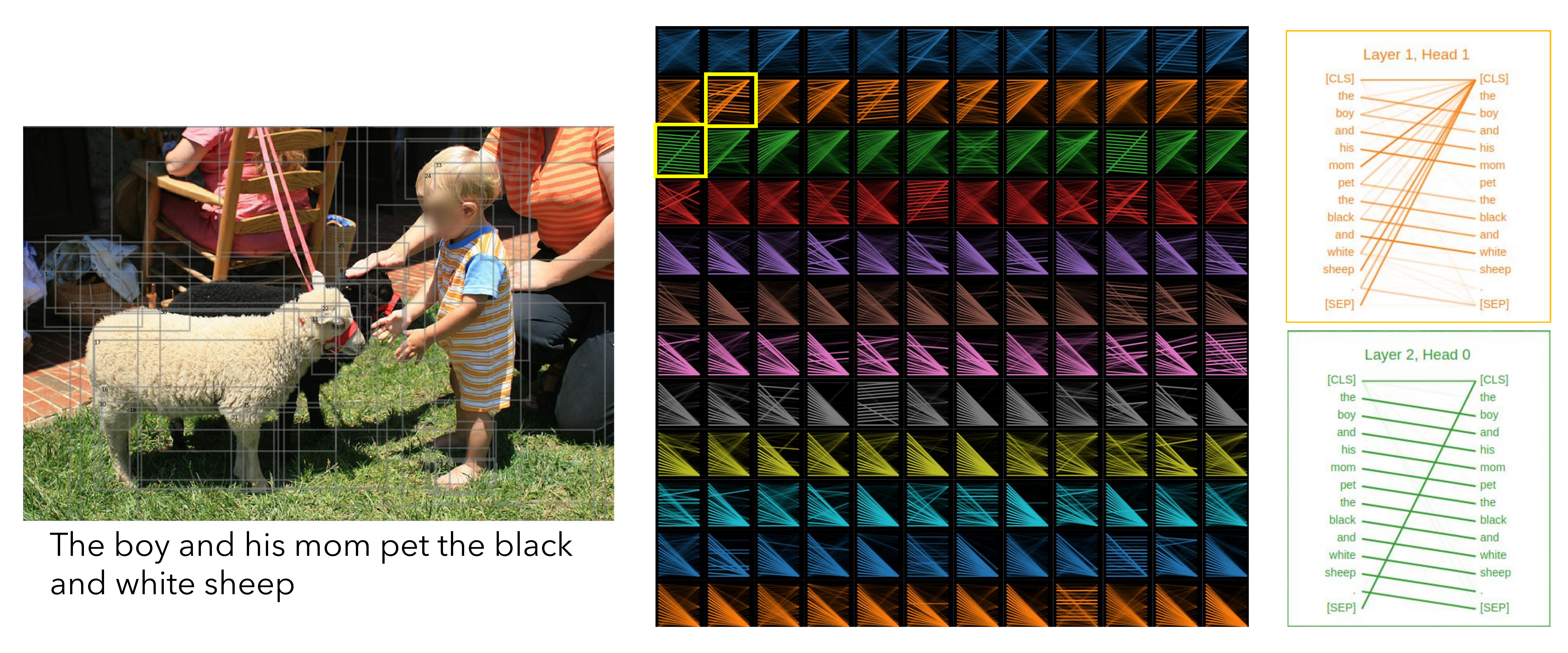}
\vspace{3pt}
\caption{Visualizations of the attentions of the pretrained
model on conceptual caption dataset using BertVis toolbox.
From left to right: Image and associate caption, sentence to sentence self-attention for all layers and all heads, sentence to sentence self-attention for Layer 1 Head 1 and Layer 2 Head 0. Our model learns the previous words attention pattern, bag of words attention pattern and next words attention pattern.}
\vspace{-10 pt}
\label{fig:attention_1}
\end{figure*}

\csubsection{Attention Visualizations}
\label{sec:attention_1}

In this section we examine the visual groundings learned by the techniques we presented in Sec.~\ref{sec:improve-vilbert}. We verify this by visualizing the attentions of our pretrained model, which is trained on the Conceptual Caption dataset. Given a test image, and corresponding caption ``The boy and his mom pet the black and white sheep'', we feed the image-caption pair as input and take the image to question co-attention for visualization. For each image patch, we use the most attended word to represent its semantic meaning, and show the patches corresponding to the visual words (`boy', `mom', `pet', `white', `sheep'). Fig.~\ref{fig:attention_0} shows the correspondence between attended regions and underlined words. We can see that the pretrained model learns meaningful visual grounding for the concept `boy', `sheep', `white' and `pet'. 

\begin{figure*}[ht]
\centering
\includegraphics[width=0.99\textwidth]{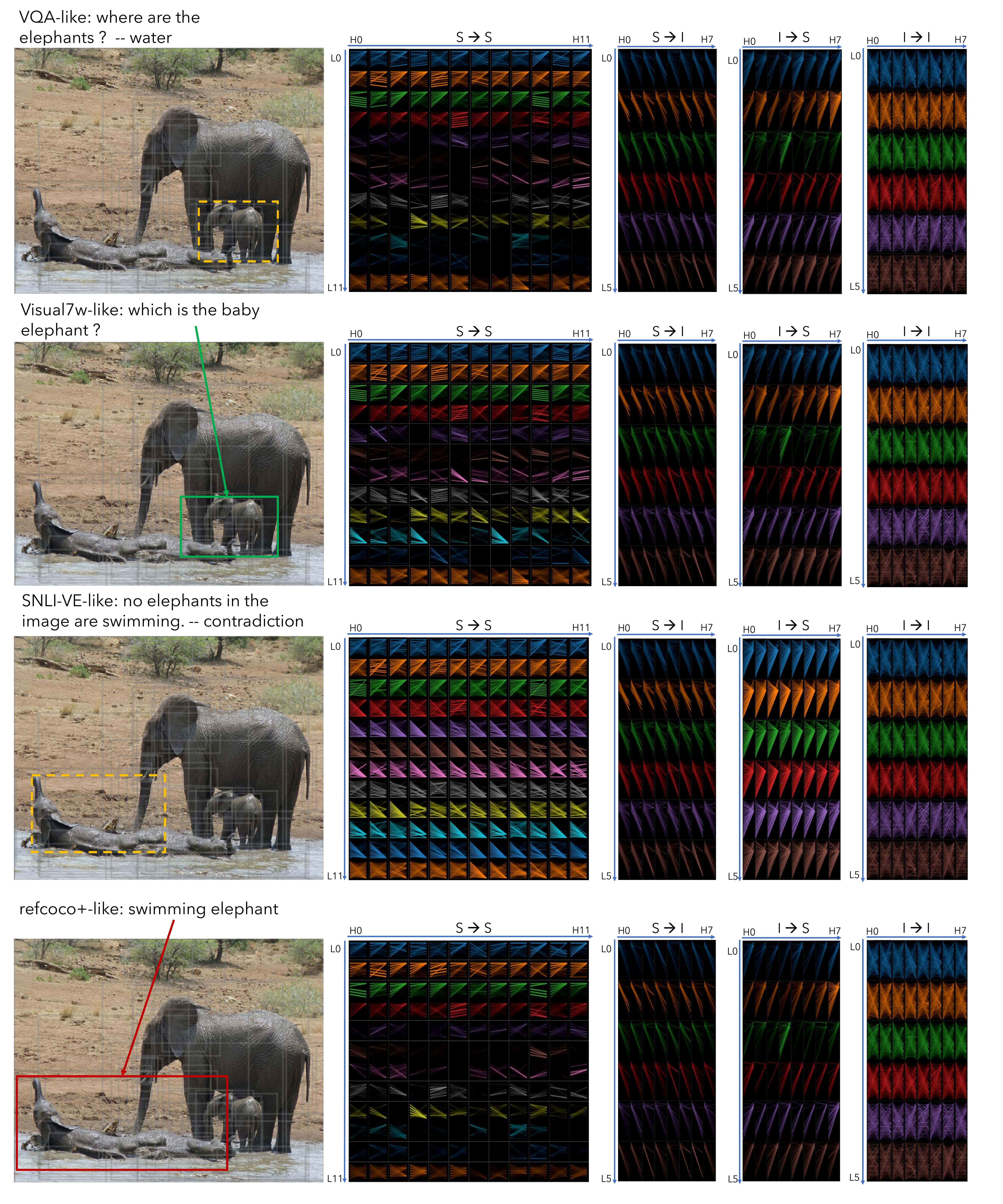}
\vspace{3pt}
\caption{Visualizations of the attentions of Our$_\texttt{AT}$ model using BertVis toolbox on each tasks. From left to right are image and associate sentence, sentence to sentence self-attention, sentence to image co-attention image to sentence co-attention image to image self-attention. Dashed orange bounding boxes in the image are the referring expression outputs regardless of tasks. The model learns to use significant different sentence to sentence self-attention pattern for different tasks.}
\vspace{-5 pt}
\label{fig:attention_2}
\end{figure*}

\begin{figure*}[ht]
\centering
\includegraphics[width=0.99\textwidth]{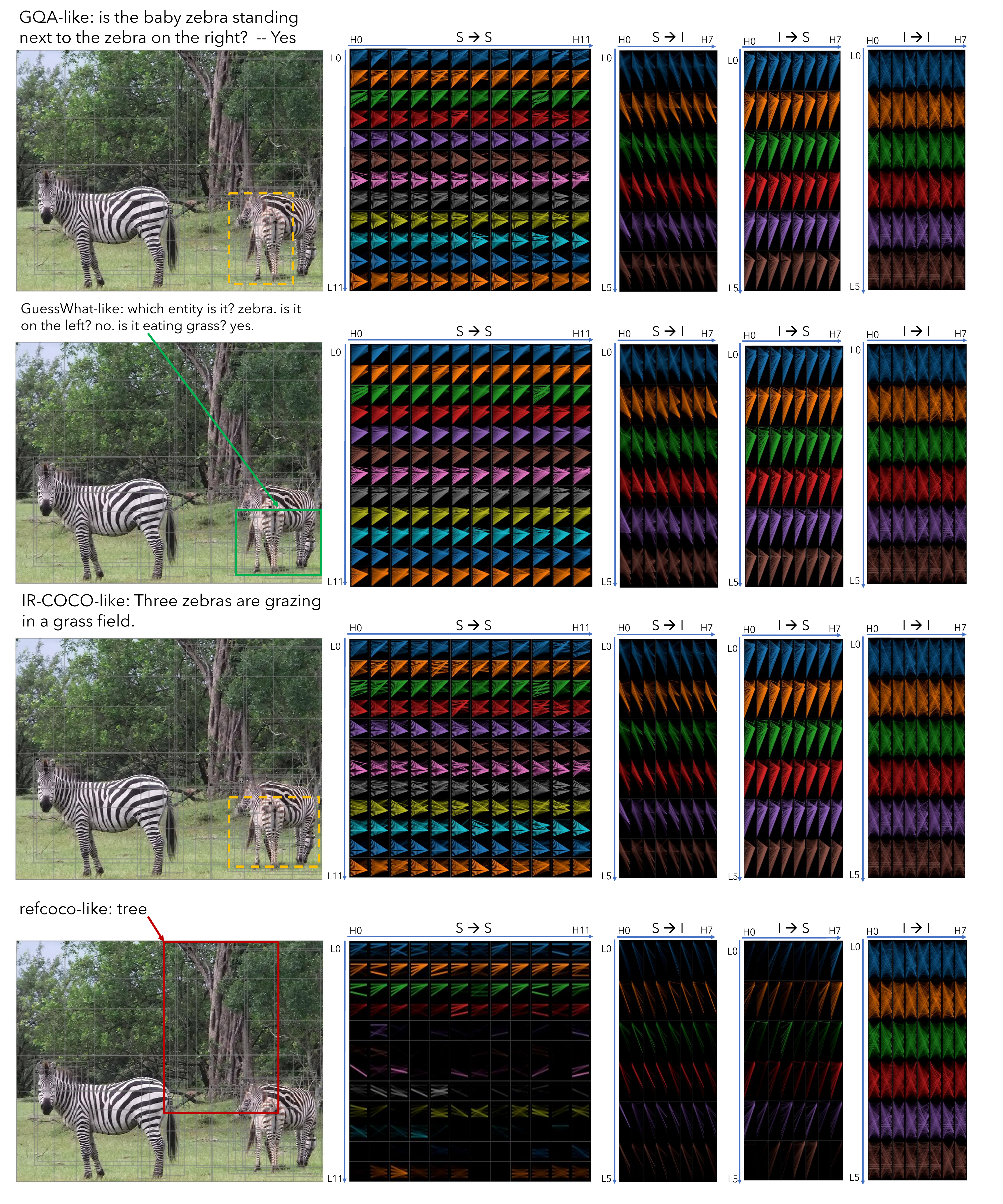}
\vspace{3pt}
\caption{Visualizations of the attentions of Our$_\texttt{AT}$ model using BertVis toolbox on each tasks. From left to right are image and associate sentence, sentence to sentence self-attention, sentence to image co-attention image to sentence co-attention image to image self-attention. Dashed orange bounding boxes in the image are the referring expression outputs regardless of tasks. The model learns to use significant different sentence to sentence self-attention pattern for different tasks. 
}
\vspace{-5 pt}
\label{fig:attention_3}
\end{figure*}

To visualize the attention for our multi-task trained model (Ours$_{\texttt{AT}}$), we use BertVis\footnote{https://github.com/jessevig/bertviz} to visualization the attention distribution on the sentence to sentence self-attention $S{\rightarrow}S$, sentence to image co-attention $S{\rightarrow}I$, image to sentence  co-attention $I{\rightarrow}S$ and image to image  self attention $I{\rightarrow}I$. 
{Fig.~\ref{fig:attention_1} shows an example of the sentence to sentence attention for all layers and all heads (middle) and a specific layer and head (right). 
We can see that our model learns the previous words attention pattern, bag of words attention pattern (Layer 1, Head 1) and next words attention pattern (Layer 2, Head 0). This shows that the model is able to generate position-aware queries and keys to calculate the attentions. 
% \marcus{why is that very insightful?}. 
To get a sense of the difference of attention distribution across different tasks, }Fig.~\ref{fig:attention_2} and Fig.~\ref{fig:attention_3} show the attention distribution on the examples of Fig.~\ref{fig:qual}. 
% \marcus{I would repeat this observation in the caption!}. 
We can see for different tasks, the model learns to use significant different sentence to sentence self-attention pattern.
% For more interesting attention visualization \marcus{what do you mean by "more interesting" here? do you want to say there is more which we are not showing...}, we will release our visualization toolbox and code for future study \marcus{this sounds like a contribution, but the captions suggest it is "just" "using BertVis"}. 

\end{document}